\documentclass{arxivtmpl}

\let\templatebibliographystyle\bibliographystyle
\renewcommand{\bibliographystyle}[1]{}

\def\1{\bm{1}}

\DeclareMathAlphabet{\mathsfit}{\encodingdefault}{\sfdefault}{m}{sl}
\SetMathAlphabet{\mathsfit}{bold}{\encodingdefault}{\sfdefault}{bx}{n}

\tcbuselibrary{skins,breakable}
\newtcolorbox{findingbox}{enhanced,breakable,frame hidden,colback=gred!6,borderline west={3pt}{0pt}{gred},sharp corners,boxsep=0pt,left=12pt,right=10pt,top=8pt,bottom=8pt,before skip=12pt,after skip=12pt}
\newtcolorbox{notebox}{enhanced,breakable,frame hidden,colback=gblue!6,borderline west={3pt}{0pt}{gblue},sharp corners,boxsep=0pt,left=12pt,right=10pt,top=8pt,bottom=8pt,before skip=12pt,after skip=12pt}
\newtcolorbox{tipbox}{enhanced,breakable,frame hidden,colback=ggreen!6,borderline west={3pt}{0pt}{ggreen},sharp corners,boxsep=0pt,left=12pt,right=10pt,top=8pt,bottom=8pt,before skip=12pt,after skip=12pt}
\newtcolorbox{warnbox}{enhanced,breakable,frame hidden,colback=gyellow!6,borderline west={3pt}{0pt}{gyellow},sharp corners,boxsep=0pt,left=12pt,right=10pt,top=8pt,bottom=8pt,before skip=12pt,after skip=12pt}

\tcbuselibrary{listings}
\newcommand{\boxfont}{\fontsize{9}{11}\selectfont}
\newtcblisting{promptbox}[1][]{%
  enhanced, listing only, breakable, frame hidden, width=\linewidth,
  colback=gblue!6, borderline west={3pt}{0pt}{gblue},
  sharp corners, boxsep=0pt,
  left=12pt, right=10pt, top=8pt, bottom=8pt, before skip=12pt, after skip=12pt,
  before upper={\def\promptlabel{#1}\ifx\promptlabel\empty\else{\sffamily\bfseries\color{gblue}#1}\par\smallskip\fi},   
  listing options={basicstyle=\ttfamily\boxfont, escapeinside={(*}{*)}},
}

\newtcolorbox{transcriptbox}{%
  enhanced, breakable, frame hidden,
  colback=ggreen!6, borderline west={3pt}{0pt}{ggreen},
  sharp corners, boxsep=0pt, fontupper=\ttfamily\boxfont, halign upper=flush left,   
  left=12pt, right=10pt, top=6pt, bottom=8pt, before skip=12pt, after skip=12pt,
}

\colorlet{codebg}{gyellow!6}
\setminted{fontsize=\boxfont, breaklines, autogobble, tabsize=4, style=vs}
\newminted[codebox]{python}{bgcolor=codebg, frame=leftline, framerule=2.5pt, rulecolor=gyellow, framesep=2.5mm}

\hypersetup{
    colorlinks=true,
    linkcolor=linkcol,
    citecolor=linkcol,
    urlcolor=linkcol,
    pdfborderstyle={},
    pdfborder={0 0 0}
}
\renewcommand\Affilfont{\rmfamily\mdseries\fontsize{11.5}{14}\selectfont}

\colorlet{perfhue}{gblue}
\colorlet{perf0}{white}
\colorlet{perf10}{perfhue!12}
\colorlet{perf20}{perfhue!22}
\colorlet{perf30}{perfhue!33}
\colorlet{perf40}{perfhue!44}
\colorlet{perf50}{perfhue!55}
\colorlet{perf60}{perfhue!66}
\colorlet{perf70}{perfhue!77}
\colorlet{perf80}{perfhue!88}

\newcommand{\RN}[1]{%
	\textup{\lowercase\expandafter{\it \romannumeral#1}}%
}

\SetKwInput{KwInput}{Input}
\SetKwInput{KwOutput}{Output}
\DontPrintSemicolon

\SetKwComment{Comment}{\color{ggreen}\# }{}

\SetKwProg{Function}{def}{:}{}

\SetKwProg{For}{for}{:}{}
\SetKwProg{If}{if}{:}{}

\let\bibliographystyle\templatebibliographystyle
\usepackage[T1]{fontenc}
\usepackage{url}
\usepackage{nicefrac}
\usepackage{pgfplots}
\pgfplotsset{compat=1.17}
\usepgfplotslibrary{groupplots}
\usetikzlibrary{arrows.meta,shapes.geometric}
\newcommand{\codedatanotetext}{}
\newcommand{\codedatanote}[1]{\gdef\codedatanotetext{#1}}
\renewcommand{\footnoterule}{\kern-3pt\hrule width\columnwidth\kern2.6pt}

\newif\ifshowrevisions
\showrevisionsfalse
\definecolor{revcolor}{RGB}{0,84,180}
\ifshowrevisions

  \DeclareRobustCommand{\revtodo}[1]{{\color{red}\textbf{[TODO: #1]}}}
\else

  \DeclareRobustCommand{\revtodo}[1]{}
\fi

\newcommand{\iptc}{\textsc{I-PTC}}
\newcommand{\ptcb}{\textsc{PTC-Bench}}
\newcommand{\ptc}{\textsc{PTC}}

\newcommand{\arxivlayout}{}

\titlespacing*{\section}{0pt}{1.1\baselineskip}{0.35\baselineskip}
\titlespacing*{\subsection}{0pt}{0.85\baselineskip}{0.15\baselineskip}
\titleformat{\section}{\rmfamily\bfseries\Large}{\thesection}{0.7em}{}
\titleformat{\subsection}{\rmfamily\bfseries\large}{\thesubsection}{0.7em}{}
\setlist[itemize]{topsep=4pt,itemsep=3pt,parsep=0pt}
\RecustomVerbatimEnvironment{verbatim}{Verbatim}{fontsize=\small,samepage=true}
\BeforeBeginEnvironment{verbatim}{\par\smallskip\noindent\begin{minipage}{\linewidth}}
\AfterEndEnvironment{verbatim}{\end{minipage}\par\smallskip}

\renewcommand{\arraystretch}{1.08}

\fancypagestyle{fancy}{%
  \fancyhf{}%
  \fancyhead[C]{\fontsize{9}{11}\selectfont\color{black!50}%
    I-PTC: Interactive Programmatic Tool Calling for Stateful Tool-Augmented Agents}%
  \fancyfoot[C]{\small\thepage}%
  }
\fancypagestyle{firstpagestyle}{%
  \fancyhf{}%
  \fancyfoot[L]{\codedatanotetext}%
  }

\renewcommand{\titlefont}{\normalfont\bfseries\fontsize{22}{26}\selectfont}

\renewcommand\Affilfont{\rmfamily\mdseries\fontsize{10.5}{13}\selectfont}

\makeatletter
\renewcommand\AB@authnote[1]{\expandafter\ifblank\expandafter{#1}{}{\,\textsuperscript{\normalfont#1}}}
\renewcommand{\maketitle}{%
  \begingroup
  \setlength{\parindent}{0pt}%
  \vspace*{-12pt}%
  {\titlefont\@title\par}%
  \vspace{12pt}%
  {\@author\par}%
  \ifx\@authornote\@empty\else
    \vspace{5pt}%
    {\fontsize{9}{11.5}\selectfont\@authornote\par}%
  \fi
  \vspace{12pt}%
  \endgroup}
\makeatother

\renewenvironment{abstract}
  {\par\begingroup\small\noindent\textbf{Abstract}\par\nobreak\vspace{3pt}\noindent\ignorespaces}
  {\par\endgroup\vspace{5pt}}

\hypersetup{
  pdftitle={I-PTC: Interactive Programmatic Tool Calling for Stateful Tool-Augmented Agents},
  pdfauthor={Shuo Yuan, Chengkun Cao, Huanzhi Mao, Joseph E. Gonzalez},
  pdfsubject={Interactive, stateful programmatic tool calling for language-model agents},
  pdfkeywords={I-PTC, PTC-Bench, tool calling, language models, stateful agents},
  bookmarksnumbered=true}

\title{\iptc{}: Interactive Programmatic Tool Calling for Stateful Tool-Augmented Agents}

\author[$*$]{Huanzhi Mao}
\author[$*$]{Chengkun Cao}
\author[$*$]{Shuo Yuan}
\author[ ]{Joseph~E.~Gonzalez}
\makeatletter\renewcommand\AB@affilnote[1]{}\makeatother
\affil[ ]{University of California, Berkeley}
\authornote{$*$\,Equal contribution.\quad Correspondence to: \texttt{huanzhimao@berkeley.edu}.}

\begin{document}

\maketitle
\setcounter{footnote}{2} 

\thispagestyle{firstpagestyle}

\begin{abstract}
Large language models are increasingly capable of writing and executing code, suggesting a natural paradigm for tool-augmented agents: expressing tool use as programs rather than isolated calls. Programmatic tool calling (\ptc{}) captures this idea through loops, batching, abstraction, and structured computation, but existing \ptc{} approaches are largely stateless, limiting their ability to reuse intermediate results and adapt over time. This is problematic for real-world tasks, where tool environments and problem-solving processes are inherently stateful. We therefore present \textbf{\iptc{}}, an interactive framework that executes stateful \ptc{}: models issue code snippets that call tools, maintain reusable state, and evolve the tool interface through API corrections. To evaluate the effectiveness of \iptc{} and the general \ptc{} setting, we introduce \textbf{\ptcb{}}, a benchmark of long-tail, production-style user requests whose difficulty comes from large initial environments, many dependent API interactions, extended planning, and recovery across follow-ups and transient failures.
Experiments on \ptcb{} show that \iptc{} improves performance on complex multi-step tasks with large tool sets while substantially reducing token consumption over conventional tool-calling baselines and the original \ptc{} baseline, highlighting stateful programmatic tool calling as a promising direction for scalable tool-augmented agents.\codedatanote{\scriptsize Project code is available at \url{https://github.com/HuanzhiMao/I-PTC}.}

\end{abstract}

\section{Introduction}

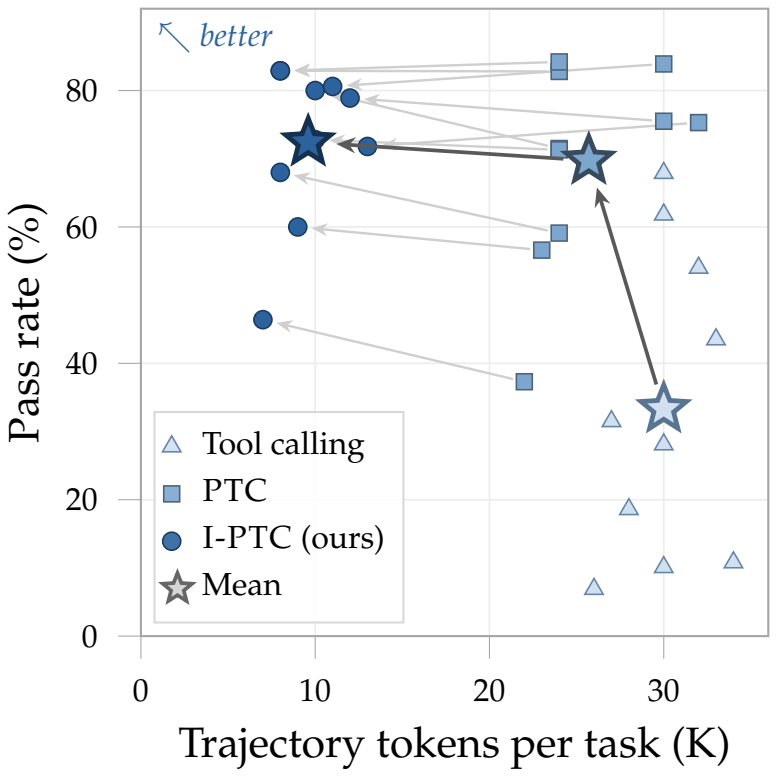
\begin{wrapfigure}{r}{0.32\textwidth}
\vspace{-1.4em}
\centering
\definecolor{toolcol}{RGB}{208,224,240}\definecolor{ptccol}{RGB}{125,166,207}\definecolor{iptccol}{RGB}{44,98,158}
\begin{tikzpicture}[font=\scriptsize]
\begin{axis}[
  width=\linewidth-28.9pt, height=\linewidth-28.9pt, scale only axis, 
  xmin=0, xmax=36, ymin=0, ymax=92,
  xtick={0,10,20,30}, ytick={0,20,40,60,80},
  xlabel={Trajectory tokens per task (K)}, ylabel={Pass rate (\%)},
  xlabel style={yshift=3pt}, ylabel style={yshift=-4pt},
  tick label style={font=\tiny}, tick align=outside, tick pos=left, axis line style={gray!70}, tick style={gray!70},
  grid=major, grid style={gray!15, line width=0.3pt},
  axis on top=false, clip=false, set layers,
  legend style={font=\tiny, at={(0.02,0.02)}, anchor=south west, draw=gray!30, fill=white, fill opacity=0.9, text opacity=1,
                inner sep=1.5pt, row sep=-1pt, column sep=2pt, legend cell align=left},
]
\node[anchor=north west, text=iptccol, font=\tiny\itshape, inner sep=2pt] at (axis cs:0.3,91.7) {$\nwarrow$ better};
\draw[gray!38, line width=0.45pt, -{Stealth[length=3pt,width=2.4pt]}, shorten <=2.2pt, shorten >=2.6pt] (axis cs:24,71.5) -- (axis cs:10,80.0);
\draw[gray!38, line width=0.45pt, -{Stealth[length=3pt,width=2.4pt]}, shorten <=2.2pt, shorten >=2.6pt] (axis cs:24,59.1) -- (axis cs:8,68.0);
\draw[gray!38, line width=0.45pt, -{Stealth[length=3pt,width=2.4pt]}, shorten <=2.2pt, shorten >=2.6pt] (axis cs:22,37.3) -- (axis cs:7,46.4);
\draw[gray!38, line width=0.45pt, -{Stealth[length=3pt,width=2.4pt]}, shorten <=2.2pt, shorten >=2.6pt] (axis cs:24,82.8) -- (axis cs:8,82.9);
\draw[gray!38, line width=0.45pt, -{Stealth[length=3pt,width=2.4pt]}, shorten <=2.2pt, shorten >=2.6pt] (axis cs:24,71.3) -- (axis cs:10,72.8);
\draw[gray!38, line width=0.45pt, -{Stealth[length=3pt,width=2.4pt]}, shorten <=2.2pt, shorten >=2.6pt] (axis cs:23,56.6) -- (axis cs:9,60.0);
\draw[gray!38, line width=0.45pt, -{Stealth[length=3pt,width=2.4pt]}, shorten <=2.2pt, shorten >=2.6pt] (axis cs:24,84.2) -- (axis cs:8,82.9);
\draw[gray!38, line width=0.45pt, -{Stealth[length=3pt,width=2.4pt]}, shorten <=2.2pt, shorten >=2.6pt] (axis cs:30,83.9) -- (axis cs:11,80.6);
\draw[gray!38, line width=0.45pt, -{Stealth[length=3pt,width=2.4pt]}, shorten <=2.2pt, shorten >=2.6pt] (axis cs:32,75.3) -- (axis cs:13,71.8);
\draw[gray!38, line width=0.45pt, -{Stealth[length=3pt,width=2.4pt]}, shorten <=2.2pt, shorten >=2.6pt] (axis cs:30,75.5) -- (axis cs:12,78.9);
\addplot[only marks, mark=triangle*, mark size=2.0pt, mark options={fill=toolcol, draw=ptccol!80!black, line width=0.3pt}] coordinates {(28,18.6) (30,10.1) (26,6.9) (30,67.9) (34,10.8) (27,31.5) (30,61.8) (32,54.0) (33,43.5) (30,28.1)};
\addlegendentry{Tool calling}
\addplot[only marks, mark=square*, mark size=1.5pt, mark options={fill=ptccol, draw=ptccol!60!black, line width=0.3pt}] coordinates {(24,71.5) (24,59.1) (22,37.3) (24,82.8) (24,71.3) (23,56.6) (24,84.2) (30,83.9) (32,75.3) (30,75.5)};
\addlegendentry{\ptc{}}
\addplot[only marks, mark=*, mark size=1.7pt, mark options={fill=iptccol, draw=iptccol!60!black, line width=0.3pt}] coordinates {(10,80.0) (8,68.0) (7,46.4) (8,82.9) (10,72.8) (9,60.0) (8,82.9) (11,80.6) (13,71.8) (12,78.9)};
\addlegendentry{\iptc{} (ours)}
\addlegendimage{legend image code/.code={\node[overlay, star, star points=5, star point ratio=2.3, minimum size=5.8pt, inner sep=0pt, line width=0.7pt, fill=gray!35, draw=black!60] at (0.26cm,0cm) {};\path (0.3cm,0cm);}}
\addlegendentry{Mean}
\begin{pgfonlayer}{axis foreground}
\draw[black!65, line width=0.7pt, -{Stealth[length=3.6pt,width=3pt]}, shorten <=4.8pt, shorten >=5.3pt] (axis cs:30.00,33.32) -- (axis cs:25.70,69.75);
\draw[black!65, line width=0.7pt, -{Stealth[length=3.6pt,width=3pt]}, shorten <=4.8pt, shorten >=5.3pt] (axis cs:25.70,69.75) -- (axis cs:9.60,72.43);
\node[star, star points=5, star point ratio=2.3, minimum size=9pt, inner sep=0pt, line width=0.9pt, fill=toolcol, draw=ptccol!70!black] at (axis cs:30.00,33.32) {};
\node[star, star points=5, star point ratio=2.3, minimum size=9pt, inner sep=0pt, line width=0.9pt, fill=ptccol, draw=ptccol!45!black] at (axis cs:25.70,69.75) {};
\node[star, star points=5, star point ratio=2.3, minimum size=9pt, inner sep=0pt, line width=0.9pt, fill=iptccol, draw=iptccol!50!black] at (axis cs:9.60,72.43) {};
\end{pgfonlayer}
\end{axis}
\end{tikzpicture}
\captionsetup{skip=0pt}\vspace{-8pt} 
\caption{\textbf{Pass rate vs.\ trajectory tokens} for ten models on \ptcb{}}
\label{fig:teaser}
\vspace{-1em}
\end{wrapfigure}

As LLMs are asked to complete increasingly complex workflows over databases, search engines, code environments, and domain-specific APIs, the key bottleneck is no longer just tool selection, but the tool-use interface itself. Conventional tool calling emits one structured call at a time and returns each result to the prompt, which works for short tasks but becomes inefficient for long workflows: each call may require another inference step, intermediate results consume context, and operations such as looping, filtering, aggregation, branching, and joining API results must be simulated through repeated natural-language reasoning. Programmatic tool calling (\ptc{})~\citep{codeact, anthropic2025advancedtooluse, letta2025programmatic} addresses this by letting agents write executable programs that orchestrate tools with loops, abstraction, conditionals, and structured computation outside the language context.

While \ptc{} improves complex tool use, standard execution remains brittle. One-shot programs must assume API behavior, return formats, and edge cases before seeing any outputs; if these assumptions are wrong, the agent may need to restart. Splitting the workflow into smaller blocks helps exploration, but loses intermediate variables and cached results, causing redundant computation and repeated tool calls. Thus, \ptc{} reduces context pollution but can limit observability and error recovery.

We therefore propose \textbf{\iptc{}}, an interactive programmatic tool-calling framework that gives the model a persistent notebook-like execution environment \citep{kluyver2016jupyter, perez2007ipython}. In \iptc{}, the model can issue code snippets, call tools, inspect intermediate variables, debug execution errors, and continue from a preserved namespace. This design retains the token-efficiency advantages of \ptc{} while recovering the observability and error-recovery benefits of conventional tool calling. By separating execution memory from language context, \iptc{} allows intermediate state to persist as executable variables rather than being repeatedly serialized into the prompt. 

Beyond within-task persistence, \iptc{} also evolves its interaction with the tool environment. \textbf{On-demand tool inspection} lets the agent dynamically acquire detailed tool information only when needed, reducing upfront prompt overhead when operating over large tool collections. \textbf{Self-evolving APIs} allow the agent to incorporate execution feedback into persistent API corrections and usage notes, so discoveries about tool behavior can be reused across later code snippets. Together, these mechanisms extend \ptc{} beyond static program execution: they turn tool discovery, execution feedback, and recurring tool-use patterns into reusable task state, improving efficiency and reliability across complex workflows.

To evaluate \iptc{} and the broader \ptc{} setting, we introduce \textbf{\ptcb{}}, a benchmark of 240 realistic user queries over simulated local APIs. \ptcb{} targets production-style workflows such as scheduling, shopping, travel planning, record analysis, personal finance review, and customer support. These tasks share a common structure: large initial environments, many dependent API calls, extended planning under constraints, and reuse of partial work across follow-ups or transient failures. Unlike existing benchmarks that focus on function-call validity, tool selection, and short-to-medium-horizon interaction, \ptcb{} provides a complementary testbed for evaluating tool-use interfaces on long-horizon, stateful workflows.

We evaluate \iptc{} against conventional tool calling and one-shot \ptc{} across \ptcb{}, MCP-Universe~\citep{luo2025mcpuniversebenchmarkinglargelanguage} and MCP-Bench~\citep{wang2025mcpbenchbenchmarkingtoolusingllm}, using both open-source and proprietary model families. 
Across these settings, \iptc{} improves task success on most evaluated models while reducing token usage and unnecessary tool calls (Figure~\ref{fig:teaser}), showing that interactive stateful execution provides a more scalable interface for complex tool-augmented agents. In summary, this paper makes the following contributions:
\begin{itemize}[leftmargin=2em]
    \item We propose \textbf{\iptc{}}, a training-free, interactive programmatic tool-calling framework that gives agents a persistent notebook-like execution environment. \iptc{} allows agents to iteratively execute code, inspect intermediate variables, debug errors, incorporate API corrections, and continue from preserved state, shifting \ptc{} from stateless execution to stateful interaction. 

    \item We introduce \textbf{\ptcb{}}, a benchmark of 240 long-tail, production-style user requests over simulated APIs, characterized by large initial environments, many dependent API interactions, extended planning under user-supplied constraints, multi-turn follow-ups, and recovery from transient API failures. The benchmark is interface-agnostic: tasks are defined by the user request and a deterministic verifier over the final answer and environment state.

    \item We empirically evaluate traditional tool calling, standard \ptc{}, and \iptc{} across \ptcb{}, MCP-Universe~\citep{luo2025mcpuniversebenchmarkinglargelanguage} and MCP-Bench~\citep{wang2025mcpbenchbenchmarkingtoolusingllm} using both open-source and proprietary model families. 
    Our results show that \iptc{} improves task success on most evaluated models while substantially reducing token consumption and redundant tool use compared with conventional tool-calling and standard \ptc{} baselines.
\end{itemize}

\section{Related Work}

\paragraph{Programmatic Tool Calling}
Recent work studies programmatic tool calling (PTC), where models write executable code to invoke tools, process results, and express control flow, rather than emitting isolated tool calls one step at a time~\citep{codeact}. This makes code a natural substrate for complex workflows involving repeated API calls, aggregation, filtering, and structured computation~\citep{anthropic2025advancedtooluse, letta2025programmatic}. Persistent Python state across turns has also been explored~\citep{qiao2023taskweaver, ran2026caveagent, zhang2025rlm, cheng2025sharingstate}. However, existing PTC approaches mostly treat code as a one-shot orchestration mechanism: the model often commits to a full program before observing tool outputs, and failures from wrong API assumptions or return formats can require restarting the workflow. \iptc{} makes PTC interactive and stateful: agents can execute partial programs, inspect intermediate outputs, reuse cached variables, and recover from errors. Unlike prior persistent-state systems, \iptc{} also treats the tool interface as adaptable state, acquiring schemas on demand and writing failure-derived corrections into tool descriptions.

\paragraph{LLM Tool Calling and Benchmarks}
Tool-using LLMs have been studied through natural-language actions, structured function calls, and API-use benchmarks, including Toolformer~\citep{toolformer}, ReAct~\citep{react}, Gorilla~\citep{gorilla}, ToolBench~\citep{toolbench}, BFCL, and $\tau$-bench~\citep{tau}. These works are central for evaluating tool selection, argument generation, function-call validity, conversational policy compliance, and short-to-medium-horizon interaction. In contrast, \ptcb{} targets a complementary regime: production-style requests with large initial states, many dependent API calls, extended planning, multi-turn reuse, and recovery from transient failures. Thus, \ptcb{} is designed to evaluate whether a tool-use interface can support long-horizon, stateful, and computation-heavy workflows, where ordinary call-observe-call interaction becomes inefficient.

\paragraph{Coding Agents}
Coding-agent systems such as SWE-agent~\citep{sweagent}, OpenHands~\citep{openhands}, and Agentless~\citep{agentless} show that LLMs can use interactive development environments to inspect state, edit code, run commands, and learn from execution feedback, with SWE-bench~\citep{swebench} serving as a benchmark for repository-level software repair. Our work transfers this insight from software engineering to general tool use: code becomes the control layer for querying APIs, transforming data, caching intermediate results, and repairing failed strategies. \iptc{} extends this setting with a persistent notebook-style environment, making programmatic tool use more adaptive than one-shot PTC and more scalable than sequential tool calling.
\section{I-PTC Framework}

\begin{figure}[t]
    \centering
    \includegraphics[width=1\textwidth]{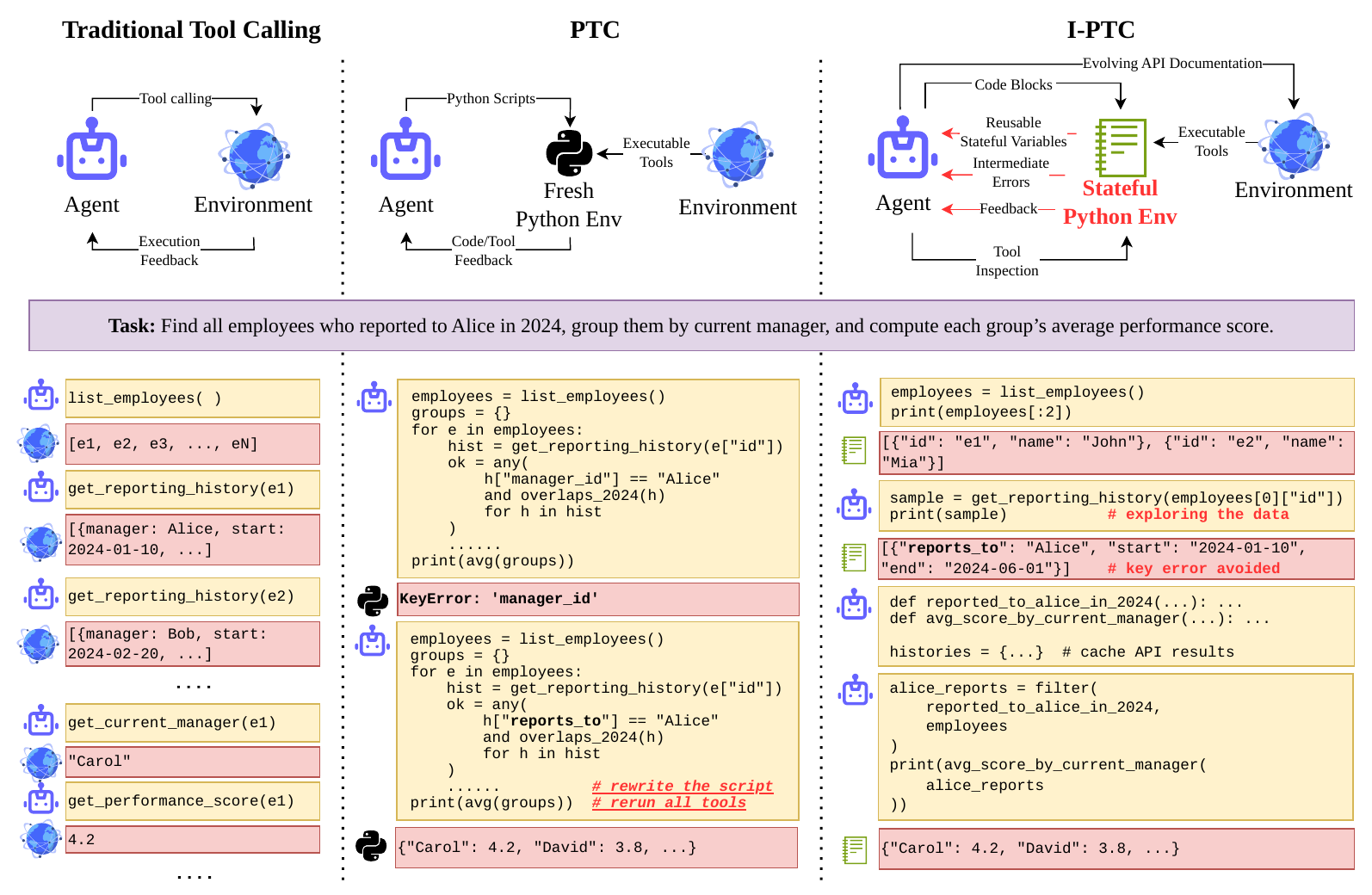}
    \caption{\textbf{Comparison of traditional tool calling, programmatic tool calling (PTC), and interactive programmatic tool calling (I-PTC).} Traditional tool calling executes tools through sequential agent-environment interactions, while PTC lets the agent generate executable code that orchestrates tool use inside a Python environment. I-PTC further extends PTC with a stateful execution environment, enabling reusable variables, intermediate-result inspection, execution feedback, tool inspection, and evolving API documentation across multiple rounds.}
    \label{fig:iptc}
\end{figure}

Although \ptc{} allows agents to express tool use as executable programs, a naive \ptc{} interface is often stateless. Each invocation executes a code snippet in a fresh sandbox and discards the namespace afterward. This design is efficient when the generated program is correct, but brittle when the agent must explore unfamiliar APIs, inspect intermediate outputs, or recover from incorrect assumptions about return formats. In such cases, the agent must either serialize intermediate state back into the prompt or re-derive it in a later call, leading to redundant computation, repeated tool calls, and limited error recovery.



\iptc{} addresses this limitation by turning one-shot program execution into an interactive, code-based tool-use interface. Tools are Python-callable functions, and by \emph{interface} we mean the protocol around execution: what API information is exposed, how schemas are acquired, what state persists, and whether failures update future tool descriptions. \iptc{} combines three distinct mechanisms: (i) \emph{persistent execution state} within a task (Section~\ref{sec:stateful_exec}), a notebook-like environment where agents write code incrementally, inspect intermediate results, recover from errors, and reuse variables across rounds, built on established infrastructure; (ii) \emph{self-evolving API documentation} across tasks (Section~\ref{sec:self_evolving}), which persists failure-derived corrections; and (iii) \emph{on-demand API discovery} (Section~\ref{sec:inspection}), which needs no persistent namespace. Together, these let API knowledge accumulate as reusable state instead of being repeatedly serialized into the context.

\subsection{Stateful Code Execution}
\label{sec:stateful_exec}


We implement stateful execution with Jupyter-notebook-cell semantics. Unlike the standard \ptc{} tool, which creates a fresh sandbox for each invocation, our interactive executor lazily initializes a single dictionary namespace on the first call and reuses it for all code cells within the same task. Variables, imports, and function definitions introduced in one cell therefore remain available in later cells. The executor runs code statement by statement, so if a cell raises an exception, state produced by earlier statements in that cell is preserved. This allows the agent to recover from mid-cell failures without discarding completed work. Although persistent notebooks can suffer from hidden or inconsistent state in human use, this issue is less problematic for agents because the interaction history is linear: each cell, output, and error is appended to the trajectory in execution order. The agent can therefore condition on the full ordered history when deciding the next action, preserving the benefits of notebook-style exploration without requiring the namespace itself to be serialized into the prompt.

\subsection{Self-Evolving APIs}
\label{sec:self_evolving}



Tool-using agents typically operate over a fixed tool surface: each API method description is hand-authored once and then treated as ground truth, even when omissions or ambiguities repeatedly cause recoverable mistakes. In \iptc{}, these mistakes naturally surface as fine-grained execution feedback because agents interact with tools through incremental code cells, probe APIs, inspect outputs, and debug errors across rounds. We treat these local failures as signals for improving the tool interface rather than merely as task-level errors.

We implement this idea with a lightweight API-documentation update mechanism. After every batch of $B$ tasks, we scan accumulated execution traces for API method errors and Python tracebacks, together with the corresponding code blocks and original method descriptions. An update agent then produces a concise one- to two-sentence addendum for each affected method. These addenda are exposed to agents in subsequent batches and persist across sessions, allowing future agents using the same APIs to benefit from previously discovered corrections, constraints, and usage notes.

\begin{figure}[t]
    \centering
    \includegraphics[width=1\textwidth]{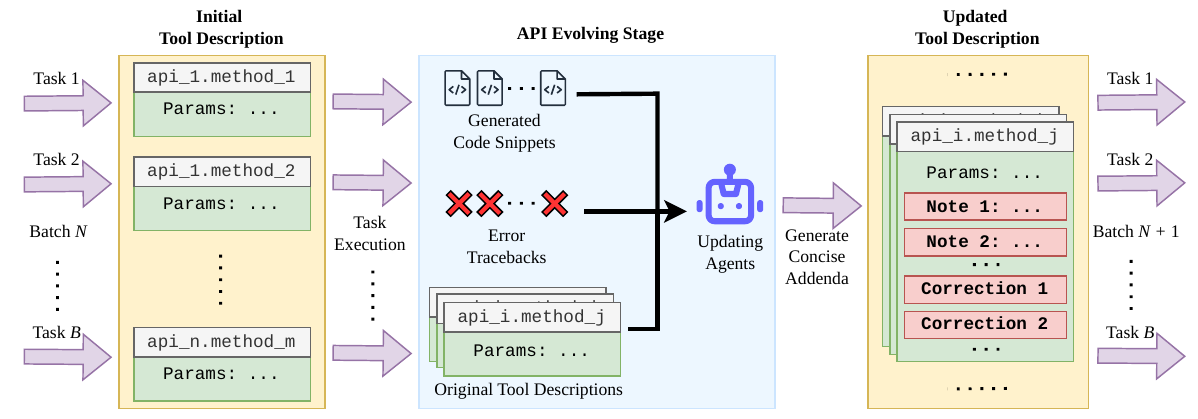}
    \caption{\textbf{Self-evolving API documentation.} After each task batch, the framework collects code snippets, API errors, tracebacks, and original tool descriptions. An updating agent converts these failure traces into concise API addenda, such as usage notes or corrections, which are appended to future tool descriptions so later agents avoid repeated mistakes.}
    \label{fig:my_pdf}
\end{figure}

\subsection{On-Demand Tool Inspection}
\label{sec:inspection}


In realistic tool-use settings, agents are often exposed to a broad catalog of APIs, even though only a small subset is relevant to any particular task. The full per-method schema for every available API is placed in the system context on every turn, causing agents to repeatedly pay for many irrelevant tool descriptions. At the same time, Python-based tool use often involves runtime uncertainty: APIs may return dictionaries, custom objects, nested records, or duck-typed objects whose useful fields and methods are not fully known until execution.

To reduce schema overhead while improving runtime observability, we integrate on-demand tool inspection into \iptc{}. Inspired by Jupyter notebook inspection syntax, \iptc{} provides a hierarchical pull-based mechanism inside code cells: a line of the form \texttt{<api>??} lists the methods available on an API server, while \texttt{<api>.<method>??} returns the full description and parameter schema for a single method. The same pattern also supports exploratory programming: after calling a tool, the agent can inspect returned objects, their types, and available fields before continuing in later cells. With inspection enabled, the initial tool list contains only the code-execution tool and bare API class names. Full method schemas enter the context only when explicitly requested, so irrelevant APIs are never inspected and never incur schema cost.
\section{Evaluation}
\label{sec:ptcbench}
To further explore \ptc{} and evaluate the effectiveness of \iptc{}, we propose \ptcb{}, a simulated-API benchmark for evaluating tool-use interfaces on complex, stateful workflows. Each task provides a natural-language user request, a fixed initial environment state, a set of available APIs, and a deterministic verifier that checks the final answer and, when applicable, the final simulated environment state.

\paragraph{Benchmark design}

\ptcb{} targets the workloads that emerge as users delegate increasingly involved tasks to LLM agents: scheduling, travel planning, and other workflow automation. Although each individual request is idiosyncratic and long-tail, these tasks share a recurring structural pattern. The user request cannot be resolved by selecting a small number of APIs. Instead, the agent is presented with a large initial environment, must issue many dependent API queries, retain and filter intermediate results, aggregate or optimize across records, and produce constraint-satisfying final actions; in multi-turn or failure-prone settings it must also reuse partial work across follow-ups or after errors. 

\begin{table}[h]
  \centering
  \small
  \caption{Complexity statistics for the PTC-Bench main subset.}
  \label{tab:ptcbench-stats}
  \begin{tabular}{lrlr}
    \toprule
    Property & Value & Property & Value \\
    \midrule
    Mean entities & 244
      & Median state size & 42 KB \\
    Mean API calls & 87.7
      & Median API calls & 51 \\
    p90 API calls & 202
      & $\geq$50 API calls & 51\% \\
    $\geq$100 API calls & 38\%
      & & \\
    \bottomrule
  \end{tabular}
\end{table}

As summarized in Table~\ref{tab:ptcbench-stats}, these properties show up quantitatively as large initial states, long tool-use trajectories, and high p90 API-call counts. The benchmark is defined entirely by these workload characteristics: tasks, initial states, and verifiers are specified at the level of the user request and the final environment, and any tool-use interface (traditional function calling, programmatic, or otherwise) can be evaluated on the same tasks. Construction, validation, and neutrality with respect to \iptc{} are detailed in Appendices~\ref{app:ptc_bench_construction}, \ref{app:validation}, and \ref{app:neutrality}.

\paragraph{Tasks and Evaluation}

\ptcb{} contains 240 tasks across three subsets, each targeting a different mode of complex tool use. \ptcb{} runs on a fully simulated local backend: each manually written task is paired with a diverse initial state, and correctness is checked deterministically by task-specific Python verifiers over the final answer and environment state. Across all subsets, tasks span realistic workflow domains such as calendar scheduling, restaurant and review analysis, ride and route optimization, grocery and delivery cost optimization, multi-API daily-life planning, etc. Example user queries are shown in Figure~\ref{fig:query_examples}. These tasks require agents to plan over multiple steps, coordinate dependent tool calls, compare alternatives, aggregate structured records, optimize under constraints, and update environment states. We organize them into the following three subsets:

\begin{itemize}[leftmargin=2em]
  \item \textbf{Main queries (100).}
  Single-turn tasks evaluating complex end-to-end tool orchestration. Each task
  gives one realistic request that may require many dependent API interactions,
  structured computation over tool outputs, and one or more valid final actions
  or answers.

  \item \textbf{Multi-turn queries (70).}
  Two-turn tasks modeling user follow-ups after an initial result. The follow-up
  depends on data and intermediate reasoning from the first turn, testing whether
  agents can reuse or recompute prior work. Accuracy is evaluated on the
  follow-up response.

  \item \textbf{Error-recovery queries (70).}
  Single-turn tasks testing robustness to mid-execution API failures. We inject
  hidden deterministic errors into selected API methods, measuring whether agents
  can inspect partial progress, recover, and continue without unnecessarily
  repeating prior work.
\end{itemize}

\begin{figure}[t]
\centering
\fontsize{7pt}{8pt}\selectfont 
\setlength{\fboxsep}{6pt}

\newcommand{\querybox}[2]{%
\fbox{%
\begin{minipage}[t][0.12\textheight][t]{\dimexpr\linewidth-2\fboxsep-2\fboxrule\relax}
\centering
\textbf{#1}
\vspace{0.3em}
\hrule
\vspace{0.5em}
\raggedright
#2
\end{minipage}%
}%
}

\begin{minipage}[t]{0.31\textwidth}
\centering
\querybox{Main Query}{
\textbf{User:} I have gift-card balances at several grocery delivery stores. 
Buy everything on my shopping list while spending as little cash out-of-pocket as possible. 
Not every store carries every item. Return JSON with \texttt{total\_out\_of\_pocket} and \texttt{assignments}.
}
\end{minipage}
\hfill
\begin{minipage}[t]{0.31\textwidth}
\centering
\querybox{Multi-turn Query}{
\textbf{Turn 1.} Schedule a 45-minute meeting with Sam, Alice, Jordan, and Alex this week. Find all UTC slots where all four are free.\\[0.35em]

\textbf{Turn 2.} Alex cannot make it. From the data already gathered, which additional slots become available with only Sam, Alice, and Jordan?
}
\end{minipage}
\hfill
\begin{minipage}[t]{0.31\textwidth}
\centering
\querybox{Error-Recovery Query}{
\textbf{User:} Find recurring payments in my Venmo history from July 2025 through March 2026.\\[0.35em]

\textbf{Injected error:} \texttt{get\_transaction\_history} returns an internal service error. Please try again.
}
\end{minipage}

\caption{\textbf{Examples of PTC-Bench query types.}
Main queries are single-shot tasks; multi-turn queries require reusing information gathered in an earlier turn; error-recovery queries inject transient tool failures that the agent must recover from. Examples are abbreviated for readability.}
\label{fig:query_examples}
\end{figure}

Across all subsets, task-specific verifiers check the final response and, when relevant, the final simulated API state, including created events, orders, or other environment updates. We report both correctness and efficiency metrics, including pass rate, token usage, and API calls.
\section{Experiments}
We evaluate all methods on \ptcb{}, the 240-task benchmark described in Section~\ref{sec:ptcbench}, including its main, multi-turn, and error-recovery subsets. 
To study whether our findings transfer beyond \ptcb{}, we also evaluate selected methods on established MCP-based tool-use benchmarks, including MCP-Universe~\citep{luo2025mcpuniversebenchmarkinglargelanguage} and MCP-Bench~\citep{wang2025mcpbenchbenchmarkingtoolusingllm}.
We evaluate a broad set of proprietary and open-weight models on \ptcb{}. 
Additional evaluation on provider-native code execution for Anthropic's built-in code execution tool for Claude and the corresponding GPT and Gemini runtimes can be found in Appendix~\ref{app:native}.

We organize evaluated methods into three interface families. \textbf{Traditional tool calling} uses the standard call-observe-call interface, where the model emits one structured tool call at a time and receives each result in the conversation context. \textbf{\ptc{}} allows the model to write executable Python code that invokes APIs and performs local computation, but each execution starts from a fresh environment, so intermediate variables and helper functions are not preserved across calls. \textbf{I-PTC} is our proposed stateful programmatic interface, where the model executes code in a persistent notebook-like environment, with a API learning batch size of $10$.

We report pass rate as the primary correctness metric. To measure efficiency, we report two complementary token metrics. \textbf{Trajectory tokens} estimate the effective token usage under caching by tokenizing each unique piece of trajectory content once, removing repeated-prefix overhead from multi-turn chat APIs. This better reflects the distinct information the agent processes and produces during the workflow. \textbf{Raw tokens} are the provider-reported input and output tokens summed across all LLM calls without accounting for caching or repeated context. Raw tokens therefore capture the uncached execution cost, while trajectory tokens approximate the practical cost when repeated context can be reused. All results are averaged over three independent end-to-end runs on the same task sets; standard deviation of the runs can be found in appendix \ref{app:std}.

\subsection{Main Results on \ptcb{}}
We first compare the three main interfaces on \ptcb{}: traditional tool calling, standard \ptc{}, and \iptc{}. Table~\ref{tab:main_results_wip} reports pass rate and token usage across model families.

\begin{table}[t]
\centering
\scriptsize
\setlength{\tabcolsep}{3.5pt}
\caption{Average accuracy and token usage across three runs on \ptcb{}. I-PTC achieves the best or near-best accuracy for most models while consistently reducing total and trajectory token usage. Standard deviations are reported in appendix \ref{app:std}.}
\label{tab:main_results_wip}
\begin{adjustbox}{max width=\textwidth}
\begin{tabular}{lccc ccc ccc}
\toprule
\multirow{2}{*}{Model}
& \multicolumn{3}{c}{Accuracy $\uparrow$}
& \multicolumn{3}{c}{Raw Tokens $\downarrow$}
& \multicolumn{3}{c}{Trajectory Tokens $\downarrow$} \\
\cmidrule(lr){2-4}
\cmidrule(lr){5-7}
\cmidrule(lr){8-10}
& Tool-only & PTC & I-PTC
& Tool-only & PTC & I-PTC
& Tool-only & PTC & I-PTC \\
\midrule

GPT 5.4
& 18.6\% & 71.5\% & \textbf{80.0\%}
& 78K & 60K & \textbf{48K}
& 28K & 24K & \textbf{10K} \\

GPT 5.4 Mini
& 10.1\% & 59.1\% & \textbf{68.0\%}
& 123K & 72K & \textbf{45K}
& 30K & 24K & \textbf{8K} \\

GPT 5.4 Nano
& 6.9\% & 37.3\% & \textbf{46.4\%}
& 85K & 57K & \textbf{53K}
& 26K & 22K & \textbf{7K} \\

\midrule
Gemini 3.1 Pro
& 67.9\% & 82.8\% & \textbf{82.9\%}
& 167K & 95K & \textbf{65K}
& 30K & 24K & \textbf{8K} \\

Gemini 3.1 Flash-Lite
& 10.8\% & 71.3\% & \textbf{72.8\%}
& 552K & 122K & \textbf{85K}
& 34K & 24K & \textbf{10K} \\

Gemini 2.5 Pro
& 31.5\% & 56.6\% & \textbf{60.0\%}
& 145K & 88K & \textbf{67K}
& 27K & 23K & \textbf{9K} \\

\midrule
Claude Opus 4.7
& 61.8\% & \textbf{84.2\%} & 82.9\%
& 172K & 184K & \textbf{85K}
& 30K & 24K & \textbf{8K} \\

Claude Sonnet 4.6
& 54.0\% & \textbf{83.9\%} & 80.6\%
& 139K & 218K & \textbf{112K}
& 32K & 30K & \textbf{11K} \\

\midrule
Qwen 3.5-397B-A17B
& 43.5\% & \textbf{75.3\%} & 71.8\%
& 232K & 180K & \textbf{118K}
& 33K & 32K & \textbf{13K} \\

Kimi K2.5
& 28.1\% & 75.5\% & \textbf{78.9\%}
& 130K & 169K & \textbf{110K}
& 30K & 30K & \textbf{12K} \\


\bottomrule
\end{tabular}
\end{adjustbox}
\end{table}

\textbf{PTC improves pass rate over traditional tool calling.} This suggests that many PTC-Bench tasks require more than selecting the right APIs: they require composing API results through structured computation. PTC supports this directly by letting the model express loops, filtering, aggregation, and branching as executable code rather than as repeated call-observe-call steps.

\textbf{I-PTC improves over standard PTC.}
I-PTC improves over standard PTC for most evaluated models, with the largest gains on GPT and consistent gains on Gemini. Claude is the main exception, where standard PTC remains slightly stronger, suggesting that some frontier models can already produce effective one-shot tool-use programs when API assumptions are correct. Overall, I-PTC offers a more robust interaction model for long-horizon tool use by letting agents inspect outputs, cache results, define helpers, and revise strategies across cells, making it especially useful when tool behavior is uncertain or recovery is needed.

\textbf{I-PTC reduces token usage.}
I-PTC reduces both raw tokens and trajectory tokens across most evaluated models. For example, Claude Opus 4.7 drops from 184K to 85K raw tokens and from 24K to 8K trajectory tokens, while Claude Sonnet 4.6 drops from 218K to 112K raw tokens and from 30K to 11K trajectory tokens. These reductions suggest that stateful execution together with tool inspection avoids repeatedly reconstructing the same workflow by allowing agents to cache intermediate results, inspect tool schema only when needed, and continue from previous cells. 

\paragraph{Stateful Execution Helps Most in Multi-Turn and Recovery Tasks.}
Table~\ref{tab:task_type_breakdown} shows that I-PTC consistently improves multi-turn and error-recovery accuracy across GPT-5.4 mini, GPT-5.4, and Gemini 3.1 Flash-Lite. These are settings where persistent state is particularly useful: I-PTC can reuse cached API results, helper functions, and intermediate computations instead of recomputing them. The Q2-only multi-turn results illustrate this benefit: second-turn API calls fall from roughly 27--30 to 4--8 across the three models. Trajectory tokens also decrease across all task subsets for all three models. Together, these results suggest that stateful execution improves efficiency broadly, with the clearest accuracy benefits on multi-turn and error-recovery tasks.

Overall, the main results show that programmatic tool calling provides a stronger interface than conventional tool calling for complex tool-use workflows, and that making this interface interactive and stateful provides additional benefits beyond stateless code execution.

\begin{table}[t]
\centering
\footnotesize
\setlength{\tabcolsep}{8pt}
\renewcommand{\arraystretch}{1.08}
\caption{\textbf{Breakdown by task type.} We compare PTC and I-PTC to isolate the effect of stateful programmatic execution. Metrics are averaged over three runs. API calls report the mean number of underlying simulated backend invocations per task. For multi-turn tasks, parentheses give API calls in the second turn (Q2) alone.}
\label{tab:task_type_breakdown}
\begin{tabular}{@{}llrrrr@{}}
\toprule
Task Type & Method & Pass $\uparrow$ & API $\downarrow$ & Raw $\downarrow$ & Traj. $\downarrow$ \\
\midrule
\multicolumn{6}{@{}l}{\textbf{GPT-5.4 mini}} \\
\addlinespace[2pt]
Multi-turn & PTC & 69.0\% & 180.9 (26.9) & 83K & 24K \\
& I-PTC & \textbf{77.1\%} & \textbf{90.6} (\textbf{4.1}) & \textbf{51K} & \textbf{9K} \\
\addlinespace[3pt]
Error-recovery & PTC & 61.3\% & 92.6 & 70K & 24K \\
& I-PTC & \textbf{72.9\%} & \textbf{74.6} & \textbf{38K} & \textbf{7K} \\
\addlinespace[3pt]
Main & PTC & 50.0\% & \textbf{69.4} & 66K & 23K \\
& I-PTC & \textbf{59.7\%} & 81.1 & \textbf{43K} & \textbf{7K} \\
\midrule
\multicolumn{6}{@{}l}{\textbf{GPT-5.4}} \\
\addlinespace[2pt]
Multi-turn & PTC & 78.1\% & 103.1 (30.6) & 75K & 24K \\
& I-PTC & \textbf{87.1\%} & \textbf{63.4 (8.0)} & \textbf{57K} & \textbf{11K} \\
\addlinespace[3pt]
Error-recovery & PTC & 78.0\% & 74.6 & 58K & 23K \\
& I-PTC & \textbf{85.7\%} & \textbf{66.6} & \textbf{42K} & \textbf{9K} \\
\addlinespace[3pt]
Main & PTC & 62.3\% & 71.8 & 51K & 24K \\
& I-PTC & \textbf{71.0\%} & \textbf{57.1} & \textbf{47K} & \textbf{10K} \\
\midrule
\multicolumn{6}{@{}l}{\textbf{Gemini 3.1 Flash-Lite}} \\
\addlinespace[2pt]
Multi-turn & PTC & 77.1\% & 250.7 (29.8) & 141K & 26K \\
& I-PTC & \textbf{78.6\%} & \textbf{102.3 (8.1)} & \textbf{93K} & \textbf{11K} \\
\addlinespace[3pt]
Error-recovery & PTC & 75.6\% & 134.3 & 110K & 24K \\
& I-PTC & \textbf{82.4\%} & \textbf{107.7} & \textbf{69K} & \textbf{9K} \\
\addlinespace[3pt]
Main & PTC & \textbf{64.1\%} & \textbf{83.9} & 116K & 23K \\
& I-PTC & 62.0\% & 119.8 & \textbf{90K} & \textbf{9K} \\
\bottomrule
\end{tabular}
\end{table}

\subsection{Generalization to Other Tool-Calling Benchmarks}

\ptcb{} is designed to stress complex programmatic tool orchestration, but a useful interface should also transfer beyond our benchmark. We therefore evaluate GPT-5.4 mini and Gemini 3.1 Flash-Lite with traditional tool calling, standard \ptc{}, and \iptc{} on two other tool use benchmarks: MCP-Universe~\citep{luo2025mcpuniversebenchmarkinglargelanguage} and MCP-Bench~\citep{wang2025mcpbenchbenchmarkingtoolusingllm}. We integrate both into the same three-mode interface used for \ptcb{}, while preserving their original tools and evaluators.

\begin{figure}[t]
\centering
\definecolor{toolcol}{RGB}{208,224,240}\definecolor{ptccol}{RGB}{125,166,207}\definecolor{iptccol}{RGB}{44,98,158}
\begin{tikzpicture}[errbar/.style={black!80, line width=0.4pt}, vlabel/.style={anchor=south, inner sep=0.6pt, font=\fontsize{5.5}{6}\selectfont}]
\begin{groupplot}[
  group style={group size=3 by 2, horizontal sep=1.05cm, vertical sep=0.32cm, xticklabels at=edge bottom},
  scale only axis, width=3.55cm, height=1.75cm, xmin=0.5, xmax=2.5, ymin=0, clip=false,
  xtick={1,2}, xticklabels={GPT-5.4 mini, Flash-Lite}, axis x line*=bottom, axis y line*=left,
  tick label style={font=\scriptsize}, label style={font=\scriptsize}, title style={font=\scriptsize\bfseries, yshift=-3pt},
  ymajorgrids, grid style={gray!20, dashed}, tick align=outside, major tick length=2pt,
  y tick label style={/pgf/number format/fixed, /pgf/number format/precision=1},
]
\nextgroupplot[title={PTC-Bench}, ylabel={Pass rate}, ymax=1.0, ytick={0,0.2,0.4,0.6,0.8}, legend style={font=\scriptsize, draw=none, at={(1.95,1.42)}, anchor=south, legend columns=3, /tikz/every even column/.append style={column sep=6pt}}]
  \addlegendimage{area legend, fill=toolcol, draw=none}\addlegendentry{Tool-only}
  \addlegendimage{area legend, fill=ptccol, draw=none}\addlegendentry{\ptc{}}
  \addlegendimage{area legend, fill=iptccol, draw=none}\addlegendentry{\iptc{}}
  \fill[toolcol] (axis cs:0.585,0) rectangle (axis cs:0.835,0.101);
  \draw[errbar] (axis cs:0.710,0.0962) -- (axis cs:0.710,0.1058) (axis cs:0.665,0.0962) -- (axis cs:0.755,0.0962) (axis cs:0.665,0.1058) -- (axis cs:0.755,0.1058);
  \node[vlabel] at (axis cs:0.710,0.1058) {0.101};
  \fill[ptccol] (axis cs:0.875,0) rectangle (axis cs:1.125,0.591);
  \draw[errbar] (axis cs:1.000,0.5741) -- (axis cs:1.000,0.6079) (axis cs:0.955,0.5741) -- (axis cs:1.045,0.5741) (axis cs:0.955,0.6079) -- (axis cs:1.045,0.6079);
  \node[vlabel] at (axis cs:1.000,0.6079) {0.591};
  \fill[iptccol] (axis cs:1.165,0) rectangle (axis cs:1.415,0.68);
  \draw[errbar] (axis cs:1.290,0.6619) -- (axis cs:1.290,0.6981) (axis cs:1.245,0.6619) -- (axis cs:1.335,0.6619) (axis cs:1.245,0.6981) -- (axis cs:1.335,0.6981);
  \node[vlabel] at (axis cs:1.290,0.6981) {0.680};
  \fill[toolcol] (axis cs:1.585,0) rectangle (axis cs:1.835,0.108);
  \draw[errbar] (axis cs:1.710,0.1000) -- (axis cs:1.710,0.1160) (axis cs:1.665,0.1000) -- (axis cs:1.755,0.1000) (axis cs:1.665,0.1160) -- (axis cs:1.755,0.1160);
  \node[vlabel] at (axis cs:1.710,0.1160) {0.108};
  \fill[ptccol] (axis cs:1.875,0) rectangle (axis cs:2.125,0.713);
  \draw[errbar] (axis cs:2.000,0.6920) -- (axis cs:2.000,0.7340) (axis cs:1.955,0.6920) -- (axis cs:2.045,0.6920) (axis cs:1.955,0.7340) -- (axis cs:2.045,0.7340);
  \node[vlabel] at (axis cs:2.000,0.7340) {0.713};
  \fill[iptccol] (axis cs:2.165,0) rectangle (axis cs:2.415,0.728);
  \draw[errbar] (axis cs:2.290,0.7170) -- (axis cs:2.290,0.7390) (axis cs:2.245,0.7170) -- (axis cs:2.335,0.7170) (axis cs:2.245,0.7390) -- (axis cs:2.335,0.7390);
  \node[vlabel] at (axis cs:2.290,0.7390) {0.728};
\nextgroupplot[title={MCP-Universe}, ylabel={Success rate}, ymax=0.75, ytick={0,0.2,0.4,0.6}]
  \fill[toolcol] (axis cs:0.585,0) rectangle (axis cs:0.835,0.278);
  \draw[errbar] (axis cs:0.710,0.2540) -- (axis cs:0.710,0.3020) (axis cs:0.665,0.2540) -- (axis cs:0.755,0.2540) (axis cs:0.665,0.3020) -- (axis cs:0.755,0.3020);
  \node[vlabel] at (axis cs:0.710,0.3020) {0.278};
  \fill[ptccol] (axis cs:0.875,0) rectangle (axis cs:1.125,0.304);
  \draw[errbar] (axis cs:1.000,0.2763) -- (axis cs:1.000,0.3317) (axis cs:0.955,0.2763) -- (axis cs:1.045,0.2763) (axis cs:0.955,0.3317) -- (axis cs:1.045,0.3317);
  \node[vlabel] at (axis cs:1.000,0.3317) {0.304};
  \fill[iptccol] (axis cs:1.165,0) rectangle (axis cs:1.415,0.422);
  \draw[errbar] (axis cs:1.290,0.3980) -- (axis cs:1.290,0.4460) (axis cs:1.245,0.3980) -- (axis cs:1.335,0.3980) (axis cs:1.245,0.4460) -- (axis cs:1.335,0.4460);
  \node[vlabel] at (axis cs:1.290,0.4460) {0.422};
  \fill[toolcol] (axis cs:1.585,0) rectangle (axis cs:1.835,0.378);
  \draw[errbar] (axis cs:1.710,0.3450) -- (axis cs:1.710,0.4110) (axis cs:1.665,0.3450) -- (axis cs:1.755,0.3450) (axis cs:1.665,0.4110) -- (axis cs:1.755,0.4110);
  \node[vlabel] at (axis cs:1.710,0.4110) {0.378};
  \fill[ptccol] (axis cs:1.875,0) rectangle (axis cs:2.125,0.489);
  \draw[errbar] (axis cs:2.000,0.4600) -- (axis cs:2.000,0.5180) (axis cs:1.955,0.4600) -- (axis cs:2.045,0.4600) (axis cs:1.955,0.5180) -- (axis cs:2.045,0.5180);
  \node[vlabel] at (axis cs:2.000,0.5180) {0.489};
  \fill[iptccol] (axis cs:2.165,0) rectangle (axis cs:2.415,0.484);
  \draw[errbar] (axis cs:2.290,0.4510) -- (axis cs:2.290,0.5170) (axis cs:2.245,0.4510) -- (axis cs:2.335,0.4510) (axis cs:2.245,0.5170) -- (axis cs:2.335,0.5170);
  \node[vlabel] at (axis cs:2.290,0.5170) {0.484};
\nextgroupplot[title={MCP-Bench}, ylabel={Overall score}, ymax=0.65, ytick={0,0.2,0.4,0.6}]
  \fill[toolcol] (axis cs:0.585,0) rectangle (axis cs:0.835,0.42);
  \draw[errbar] (axis cs:0.710,0.4103) -- (axis cs:0.710,0.4297) (axis cs:0.665,0.4103) -- (axis cs:0.755,0.4103) (axis cs:0.665,0.4297) -- (axis cs:0.755,0.4297);
  \node[vlabel] at (axis cs:0.710,0.4297) {0.420};
  \fill[ptccol] (axis cs:0.875,0) rectangle (axis cs:1.125,0.427);
  \draw[errbar] (axis cs:1.000,0.4171) -- (axis cs:1.000,0.4369) (axis cs:0.955,0.4171) -- (axis cs:1.045,0.4171) (axis cs:0.955,0.4369) -- (axis cs:1.045,0.4369);
  \node[vlabel] at (axis cs:1.000,0.4369) {0.427};
  \fill[iptccol] (axis cs:1.165,0) rectangle (axis cs:1.415,0.439);
  \draw[errbar] (axis cs:1.290,0.4365) -- (axis cs:1.290,0.4415) (axis cs:1.245,0.4365) -- (axis cs:1.335,0.4365) (axis cs:1.245,0.4415) -- (axis cs:1.335,0.4415);
  \node[vlabel] at (axis cs:1.290,0.4415) {0.439};
  \fill[toolcol] (axis cs:1.585,0) rectangle (axis cs:1.835,0.354);
  \draw[errbar] (axis cs:1.710,0.3430) -- (axis cs:1.710,0.3650) (axis cs:1.665,0.3430) -- (axis cs:1.755,0.3430) (axis cs:1.665,0.3650) -- (axis cs:1.755,0.3650);
  \node[vlabel] at (axis cs:1.710,0.3650) {0.354};
  \fill[ptccol] (axis cs:1.875,0) rectangle (axis cs:2.125,0.385);
  \draw[errbar] (axis cs:2.000,0.3720) -- (axis cs:2.000,0.3980) (axis cs:1.955,0.3720) -- (axis cs:2.045,0.3720) (axis cs:1.955,0.3980) -- (axis cs:2.045,0.3980);
  \node[vlabel] at (axis cs:2.000,0.3980) {0.385};
  \fill[iptccol] (axis cs:2.165,0) rectangle (axis cs:2.415,0.403);
  \draw[errbar] (axis cs:2.290,0.3940) -- (axis cs:2.290,0.4120) (axis cs:2.245,0.3940) -- (axis cs:2.335,0.3940) (axis cs:2.245,0.4120) -- (axis cs:2.335,0.4120);
  \node[vlabel] at (axis cs:2.290,0.4120) {0.403};
\nextgroupplot[ymax=45, ytick={0,10,20,30,40}, ylabel={Traj. tokens (K)}]
  \fill[toolcol] (axis cs:0.585,0) rectangle (axis cs:0.835,30.5);
  \draw[errbar] (axis cs:0.710,30.3700) -- (axis cs:0.710,30.6300) (axis cs:0.665,30.3700) -- (axis cs:0.755,30.3700) (axis cs:0.665,30.6300) -- (axis cs:0.755,30.6300);
  \node[vlabel] at (axis cs:0.710,30.6300) {30.5K};
  \fill[ptccol] (axis cs:0.875,0) rectangle (axis cs:1.125,24.0);
  \draw[errbar] (axis cs:1.000,23.9200) -- (axis cs:1.000,24.0800) (axis cs:0.955,23.9200) -- (axis cs:1.045,23.9200) (axis cs:0.955,24.0800) -- (axis cs:1.045,24.0800);
  \node[vlabel] at (axis cs:1.000,24.0800) {24.0K};
  \fill[iptccol] (axis cs:1.165,0) rectangle (axis cs:1.415,7.9);
  \draw[errbar] (axis cs:1.290,7.8200) -- (axis cs:1.290,7.9800) (axis cs:1.245,7.8200) -- (axis cs:1.335,7.8200) (axis cs:1.245,7.9800) -- (axis cs:1.335,7.9800);
  \node[vlabel] at (axis cs:1.290,7.9800) {7.9K};
  \fill[toolcol] (axis cs:1.585,0) rectangle (axis cs:1.835,34.5);
  \draw[errbar] (axis cs:1.710,33.7200) -- (axis cs:1.710,35.2800) (axis cs:1.665,33.7200) -- (axis cs:1.755,33.7200) (axis cs:1.665,35.2800) -- (axis cs:1.755,35.2800);
  \node[vlabel] at (axis cs:1.710,35.2800) {34.5K};
  \fill[ptccol] (axis cs:1.875,0) rectangle (axis cs:2.125,24.5);
  \draw[errbar] (axis cs:2.000,24.0800) -- (axis cs:2.000,24.9200) (axis cs:1.955,24.0800) -- (axis cs:2.045,24.0800) (axis cs:1.955,24.9200) -- (axis cs:2.045,24.9200);
  \node[vlabel] at (axis cs:2.000,24.9200) {24.5K};
  \fill[iptccol] (axis cs:2.165,0) rectangle (axis cs:2.415,9.8);
  \draw[errbar] (axis cs:2.290,9.3700) -- (axis cs:2.290,10.2300) (axis cs:2.245,9.3700) -- (axis cs:2.335,9.3700) (axis cs:2.245,10.2300) -- (axis cs:2.335,10.2300);
  \node[vlabel] at (axis cs:2.290,10.2300) {9.8K};
\nextgroupplot[ymax=32, ytick={0,10,20,30}]
  \fill[toolcol] (axis cs:0.585,0) rectangle (axis cs:0.835,15.7);
  \draw[errbar] (axis cs:0.710,15.4800) -- (axis cs:0.710,15.9200) (axis cs:0.665,15.4800) -- (axis cs:0.755,15.4800) (axis cs:0.665,15.9200) -- (axis cs:0.755,15.9200);
  \node[vlabel] at (axis cs:0.710,15.9200) {15.7K};
  \fill[ptccol] (axis cs:0.875,0) rectangle (axis cs:1.125,16.4);
  \draw[errbar] (axis cs:1.000,15.1500) -- (axis cs:1.000,17.6500) (axis cs:0.955,15.1500) -- (axis cs:1.045,15.1500) (axis cs:0.955,17.6500) -- (axis cs:1.045,17.6500);
  \node[vlabel] at (axis cs:1.000,17.6500) {16.4K};
  \fill[iptccol] (axis cs:1.165,0) rectangle (axis cs:1.415,13.4);
  \draw[errbar] (axis cs:1.290,12.6900) -- (axis cs:1.290,14.1100) (axis cs:1.245,12.6900) -- (axis cs:1.335,12.6900) (axis cs:1.245,14.1100) -- (axis cs:1.335,14.1100);
  \node[vlabel] at (axis cs:1.290,14.1100) {13.4K};
  \fill[toolcol] (axis cs:1.585,0) rectangle (axis cs:1.835,18.4);
  \draw[errbar] (axis cs:1.710,15.3000) -- (axis cs:1.710,21.5000) (axis cs:1.665,15.3000) -- (axis cs:1.755,15.3000) (axis cs:1.665,21.5000) -- (axis cs:1.755,21.5000);
  \node[vlabel] at (axis cs:1.710,21.5000) {18.4K};
  \fill[ptccol] (axis cs:1.875,0) rectangle (axis cs:2.125,23.2);
  \draw[errbar] (axis cs:2.000,21.1000) -- (axis cs:2.000,25.3000) (axis cs:1.955,21.1000) -- (axis cs:2.045,21.1000) (axis cs:1.955,25.3000) -- (axis cs:2.045,25.3000);
  \node[vlabel] at (axis cs:2.000,25.3000) {23.2K};
  \fill[iptccol] (axis cs:2.165,0) rectangle (axis cs:2.415,22.4);
  \draw[errbar] (axis cs:2.290,20.9000) -- (axis cs:2.290,23.9000) (axis cs:2.245,20.9000) -- (axis cs:2.335,20.9000) (axis cs:2.245,23.9000) -- (axis cs:2.335,23.9000);
  \node[vlabel] at (axis cs:2.290,23.9000) {22.4K};
\nextgroupplot[ymax=200, ytick={0,50,100,150}]
  \fill[toolcol] (axis cs:0.585,0) rectangle (axis cs:0.835,110.9);
  \draw[errbar] (axis cs:0.710,68.2000) -- (axis cs:0.710,153.6000) (axis cs:0.665,68.2000) -- (axis cs:0.755,68.2000) (axis cs:0.665,153.6000) -- (axis cs:0.755,153.6000);
  \node[vlabel] at (axis cs:0.710,153.6000) {110.9K};
  \fill[ptccol] (axis cs:0.875,0) rectangle (axis cs:1.125,68.6);
  \draw[errbar] (axis cs:1.000,18.5000) -- (axis cs:1.000,118.7000) (axis cs:0.955,18.5000) -- (axis cs:1.045,18.5000) (axis cs:0.955,118.7000) -- (axis cs:1.045,118.7000);
  \node[vlabel] at (axis cs:1.000,118.7000) {68.6K};
  \fill[iptccol] (axis cs:1.165,0) rectangle (axis cs:1.415,22.1);
  \draw[errbar] (axis cs:1.290,21.7100) -- (axis cs:1.290,22.4900) (axis cs:1.245,21.7100) -- (axis cs:1.335,21.7100) (axis cs:1.245,22.4900) -- (axis cs:1.335,22.4900);
  \node[vlabel] at (axis cs:1.290,22.4900) {22.1K};
  \fill[toolcol] (axis cs:1.585,0) rectangle (axis cs:1.835,84.9);
  \draw[errbar] (axis cs:1.710,76.4000) -- (axis cs:1.710,93.4000) (axis cs:1.665,76.4000) -- (axis cs:1.755,76.4000) (axis cs:1.665,93.4000) -- (axis cs:1.755,93.4000);
  \node[vlabel] at (axis cs:1.710,93.4000) {84.9K};
  \fill[ptccol] (axis cs:1.875,0) rectangle (axis cs:2.125,37.5);
  \draw[errbar] (axis cs:2.000,22.5000) -- (axis cs:2.000,52.5000) (axis cs:1.955,22.5000) -- (axis cs:2.045,22.5000) (axis cs:1.955,52.5000) -- (axis cs:2.045,52.5000);
  \node[vlabel] at (axis cs:2.000,52.5000) {37.5K};
  \fill[iptccol] (axis cs:2.165,0) rectangle (axis cs:2.415,28.0);
  \draw[errbar] (axis cs:2.290,23.1000) -- (axis cs:2.290,32.9000) (axis cs:2.245,23.1000) -- (axis cs:2.335,23.1000) (axis cs:2.245,32.9000) -- (axis cs:2.335,32.9000);
  \node[vlabel] at (axis cs:2.290,32.9000) {28.0K};
\end{groupplot}
\end{tikzpicture}
\caption{Generalization to existing tool-calling benchmarks. We evaluate GPT-5.4 mini and Gemini 3.1 Flash-Lite with traditional tool calling, standard \ptc{}, and \iptc{} on \ptcb{}, MCP-Universe subset and MCP-Bench subset.}
\label{fig:existing_benchmark_generalization}
\end{figure}
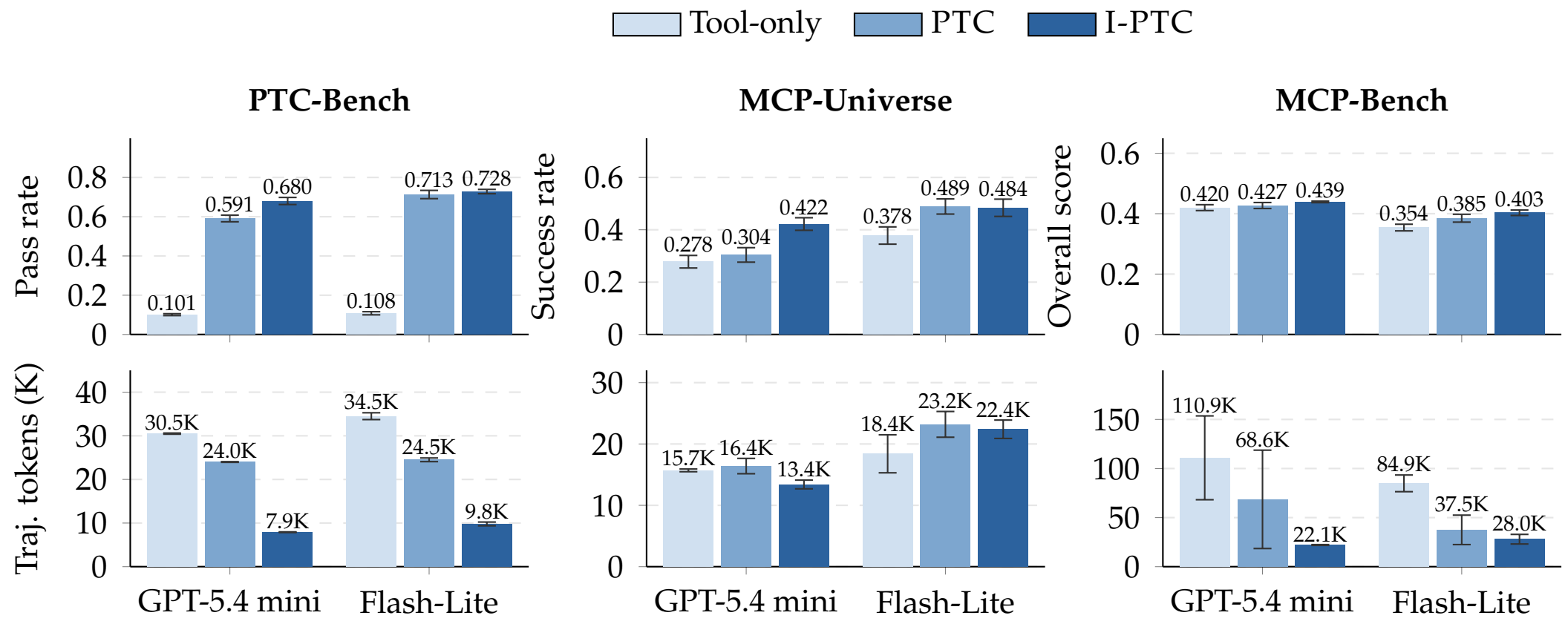

Figure~\ref{fig:existing_benchmark_generalization} shows that \iptc{} transfers to MCP-based benchmarks across models: with GPT-5.4 mini, it outperforms traditional tool calling and standard \ptc{} while using fewer tokens; with Gemini 3.1 Flash-Lite, it outperforms \ptc{} on MCP-Bench and ties it on MCP-Universe. These results suggest that the benefits of stateful programmatic execution extend beyond \ptcb{} to broader tool-use settings.

\subsection{Execution Overhead}
\label{sec:overhead}
Token usage alone does not capture net efficiency, so we instrumented the executor to record wall-clock time, code-execution time, and namespace size. The persistent environment adds little overhead: the median \iptc{} cell runs in 5--9\,ms across models (Table~\ref{tab:overhead_wall}), and in an instrumented GPT-5.4 mini run (Table~\ref{tab:overhead_instr}), \iptc{} spends only 0.2\% of its wall time executing code with a peak namespace of 0.44\,MB, whereas stateless \ptc{} spends 38.7\% because a few one-shot programs committed to expensive brute-force strategies before observing intermediate outputs. In end-to-end wall time (Table~\ref{tab:overhead_wall}), \iptc{} is faster than \ptc{} for three of the four models and slower only for \mbox{Gemini} 3.1 Pro (53.7\,s vs.\ 44.6\,s per task). Since cells take milliseconds, these differences come from provider latency and the number of model rounds rather than from notebook persistence.

\begin{table}[t]
\centering
\scriptsize
\caption{\textbf{Execution overhead.} Med.\ cell is the median execution time of \iptc{}.}
\label{tab:overhead}
\captionsetup[sub]{skip=2pt}
\begin{subtable}[t]{0.54\textwidth}
\centering
\caption{Mean wall time per task over \ptcb{} tasks.}
\label{tab:overhead_wall}
\begin{tabular}[t]{lcccc}
\toprule
Model & Tool-only & \ptc{} & \iptc{} & Med.\ cell \\
\midrule
GPT-5.4 mini & \textbf{14.0\,s} & 19.1\,s & 15.5\,s & 8\,ms \\
Gemini 3.1 Pro & 117.8\,s & \textbf{44.6\,s} & 53.7\,s & 5\,ms \\
Claude Sonnet 4.6 & 84.9\,s & 96.5\,s & \textbf{65.6\,s} & 9\,ms \\
Kimi K2.5 & 239.9\,s & 179.9\,s & \textbf{147.4\,s} & 8\,ms \\
\bottomrule
\end{tabular}
\end{subtable}\hfill
\begin{subtable}[t]{0.44\textwidth}
\centering
\setlength{\tabcolsep}{3pt}
\caption{GPT-5.4 mini, main subset.}
\label{tab:overhead_instr}
\begin{tabular}[t]{lccc}
\toprule
Condition & Total exec. & Exec. share & Peak namespace \\
\midrule
Tool-only & 0\,s & 0.0\% & -- \\
\ptc{} & 747.3\,s & 38.7\% & 0.53\,MB \\
\iptc{} & 1.9\,s & 0.2\% & 0.44\,MB \\
\bottomrule
\end{tabular}
\end{subtable}
\end{table}

\subsection{Ablation Study}
\label{sec:ablation}
We ablate \iptc{} to identify which components drive the gains. Specifically, we ask: (1) how much benefit comes from the stateful cell interface, and whether agents reuse variables and helper functions across cells and (2) whether self-evolving API documentation reduces repeated API misuse. These ablations separate the effects of persistent execution state and learned tool-surface improvements.

\begin{table}[t]
\centering
\scriptsize
\caption{\textbf{Component ablation} on GPT-5.4 mini and Gemini 3.1 Flash-Lite. On-demand inspection and self-evolving documentation are added to persistent cells (upper block) and to stateless execution (lower block); mean $\pm$ std over three runs.}
\label{tab:component_ablation}
\setlength{\tabcolsep}{3.5pt}
\begin{adjustbox}{max width=\textwidth}
\begin{tabular}{l ccc ccc}
\toprule
& \multicolumn{3}{c}{GPT-5.4 mini} & \multicolumn{3}{c}{Gemini 3.1 Flash-Lite} \\
\cmidrule(lr){2-4}\cmidrule(lr){5-7}
Configuration & Pass $\uparrow$ & Raw Tok. $\downarrow$ & Traj. Tok. $\downarrow$ & Pass $\uparrow$ & Raw Tok. $\downarrow$ & Traj. Tok. $\downarrow$ \\
\midrule
\multicolumn{7}{l}{\textit{Persistent cells}} \\
Raw \iptc{} & $61.3\pm0.8$ & 129K$\pm$5K & 25.7K$\pm$0.2K & $71.9\pm2.7$ & 130K$\pm$2K & 24.1K$\pm$0.1K \\
Raw \iptc{} + inspection & $\mathbf{68.1\pm1.7}$ & \underline{44K$\pm$1K} & \textbf{6.8K$\pm$0.1K} & $\underline{72.8\pm2.8}$ & \textbf{84K$\pm$5K} & \textbf{9.3K$\pm$0.4K} \\
Raw \iptc{} + documentation & $62.9\pm3.3$ & 124K$\pm$0.5K & 24.8K$\pm$0.2K & $70.1\pm0.9$ & 128K$\pm$4K & 23.9K$\pm$0.2K \\
Full \iptc{} & $\underline{68.0\pm1.8}$ & 45K$\pm$0.4K & \underline{7.9K$\pm$0.1K} & $\mathbf{72.8\pm1.1}$ & \underline{85K$\pm$5K} & \underline{9.8K$\pm$0.4K} \\
\midrule
\multicolumn{7}{l}{\textit{Stateless execution}} \\
PTC & $59.1\pm1.7$ & 72K$\pm$2K & 24.0K$\pm$0.1K & $71.3\pm2.1$ & 122K$\pm$6K & 24.5K$\pm$0.4K \\
PTC + inspection & $65.8\pm0.7$ & \textbf{40K$\pm$1K} & 8.8K$\pm$0.4K & $70.1\pm0.5$ & 86K$\pm$2K & 10.2K$\pm$0.3K \\
PTC + documentation & $57.2\pm1.7$ & 63K$\pm$2K & 23.0K$\pm$0.2K & $66.8\pm1.7$ & 126K$\pm$1K & 24.2K$\pm$0.1K \\
PTC + inspection + documentation & $64.7\pm4.0$ & 44K$\pm$3K & 9.4K$\pm$0.4K & $70.7\pm0.6$ & 115K$\pm$7K & 12.2K$\pm$0.3K \\
\bottomrule
\end{tabular}
\end{adjustbox}
\end{table}

\paragraph{Stateful cell execution and state reuse.}
We first isolate I-PTC's stateful execution environment.  Table~\ref{tab:component_ablation} compares persistent cells with stateless execution on GPT-5.4 mini and Gemini 3.1 Flash-Lite, with and without on-demand inspection and documentation. Persistent cells raise the mean pass rate in all eight matched configurations, 
by 0.6--5.7 points, 
\begin{wraptable}{r}{0.4\textwidth}
\centering
\scriptsize
\caption{\textbf{Cross-cell state reuse} (GPT-5.4, three runs). Cells that fail when rerun in a fresh namespace count as state-reusing.}
\label{tab:state_reuse}
\begin{tabular}{lrrr}
\toprule
Subset & Total & Reuse & Rate \\
\midrule
Main & 279.3 & 158.7 & \textbf{56.8\%} \\
Multi-turn & 195.0 & 116.3 & \textbf{59.7\%} \\
Error-recovery & 238.3 & 160.0 & \textbf{67.1\%} \\
Overall & 712.7 & 435.0 & \textbf{61.0\%} \\
\bottomrule
\end{tabular}
\vspace{-0.8em}
\end{wraptable}
although several of these gains are within run-to-run variation. 
On-demand inspection accounts for most of the token savings, with or without a persistent namespace.
Table~\ref{tab:state_reuse} verifies that agents use this persistence. We rerun each cell in a fresh namespace and count it as state-reusing if it fails due to missing variables, imports, or helper definitions. Overall, $61.0\%$ of cells depend on prior states, with especially high reuse on error-recovery tasks. Thus, I-PTC does not simply split one-shot programs into smaller snippets but reuses intermediate states across the trajectory.

\paragraph{Self-evolving API documentation.}
We also evaluate whether I-PTC's self-evolving API documentation reduces repeated tool misuse. The agent accumulates lightweight corrections from API failures and applies them every $B$ examples, where smaller $B$ means more frequent updates and $B=100$ means effectively no update within each task type.

\begin{figure}[t]
\centering
\hspace*{-0.03\textwidth}
\begin{minipage}[t]{0.49\textwidth}
\vspace{0pt}
\centering
\scriptsize
\textbf{(a) Documentation-update ablation}\\[0.5em]
\renewcommand{\arraystretch}{1.15}
\begin{tabular}{lrrr}
\toprule
Batch Size & API Err. $\downarrow$ & Pass $\uparrow$ & Raw Tok. $\downarrow$ \\
  \midrule
  1   & 82.3  & 80.6\% & 47K \\                                  
  5   & 91.3  & 80.6\% & 48K \\                                  
  10  & 94.3  & 80.3\% & 49K \\                                  
  20  & 88.0  & 80.8\% & 47K \\                                
  50  & 148.3 & 80.8\% & 49K \\
  100 & 181.3 & 81.8\% & 49K  \\   
\bottomrule
\end{tabular}
\renewcommand{\arraystretch}{1.0}
\end{minipage}
\hspace{0.035\textwidth}
\begin{minipage}[t]{0.48\textwidth}
\vspace{0pt}
\centering
\scriptsize
\textbf{(b) API errors vs. batch size}\\[-0.2em]
\label{fig:api_error}
\includegraphics[width=0.95\linewidth]{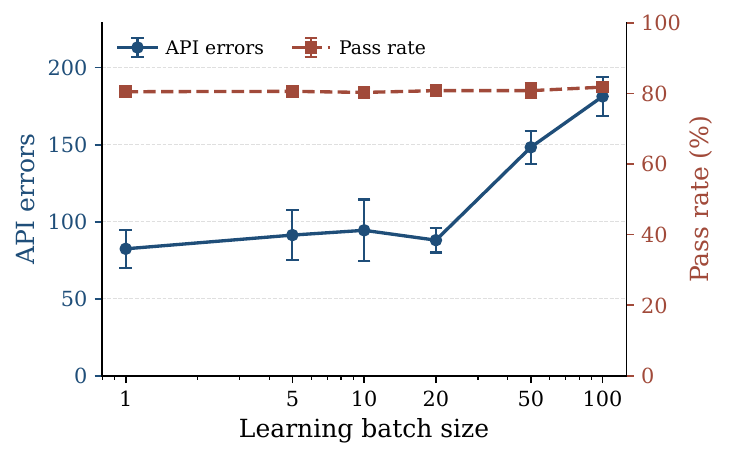}
\end{minipage}

\vspace{-0.3em}
\caption{\textbf{Effect of self-evolving API documentation.}
(a) Ablation over documentation-update batch size.
(b) API errors increase as the learning batch size grows on the 100 main queries; batch size $100$ corresponds to no effective learning during the run.}
\label{fig:doc_learner_ablation}
\end{figure}

Figure~\ref{fig:doc_learner_ablation} shows that pass rate and token cost remain stable across batch sizes, but API errors rise sharply when updates are delayed. This suggests that self-evolving documentation mainly improves tool-use reliability by preventing repeated API mistakes. While our benchmark has only 240 queries, this effect should become increasingly important as tasks, tools, and repeated API interactions scale in the real world.

\section{Limitations}
\label{sec:limitation}
Our work has three main limitations. First, self-evolving API documentation remains preliminary: although it helps reduce API errors, incorrect addenda may persist across tasks and affect later executions. Future work should validate updates before persistence and support versioning or rollback.

Second, I-PTC inherits risks and overheads from notebook-style persistent execution. Persistent state may introduce hidden-state errors from stale variables or unintended side effects. The measured overhead is small (Section~\ref{sec:overhead}), but \iptc{} is not uniformly faster than \ptc{}, and our multi-turn tasks have only two turns; longer sessions may amplify stale-state errors and namespace growth. Future work could improve persistent-state management through automatic cleanup, dependency tracking, stale-state detection, and selective reset.

Third, programmatic interfaces require sufficient model capability. In the Qwen3.5 family, they slightly underperform traditional tool calling at 0.8B and 2B parameters and help only from 4B upward (Appendix~\ref{app:scale}).

\section{Conclusion}
We introduce \iptc{}, an interactive extension of \ptc{} that moves tool use from isolated code execution to a stateful environment. Unlike standard \ptc{}, \iptc{} lets agents incrementally write code, inspect outputs, cache API results, and recover from errors. We also introduce \ptcb{}, a benchmark of long-tail, production-style user requests covering large initial environments, many dependent API interactions, extended planning, multi-turn follow-ups, and recovery from transient failures. 
Across \ptcb{}, \iptc{} improves pass rate on most evaluated models while significantly reducing token usage compared with traditional tool calling and standard \ptc{}  and redundant API calls relative to \ptc{}.
Overall, for realistic long-horizon workflows, our results suggest that future tool-augmented agents should move beyond isolated tool invocation toward persistent, stateful, and programmatic interaction environments.
\section*{Acknowledgment}
Sky Computing Lab is supported by gifts from Accenture, Algorithmic SuperIntelligence Labs, Amazon, AMD, Anyscale, Broadcom, cmpnd, Google, IBM, Intel, Intesa Sanpaolo, Lambda, \mbox{Lightspeed}, Mirendil, NVIDIA, Samsung SDS, and VESSL.

\clearpage
\begingroup
\setlength{\bibsep}{1pt plus 0.5pt}
\bibliography{references}
\endgroup

\clearpage
\beginsupplement
\crefalias{section}{appendix}
\raggedbottom
\makeatletter
\setlength{\@fptop}{0pt}
\setlength{\@fpsep}{16pt}
\setlength{\@fpbot}{0pt plus 1fil}
\makeatother

\section*{Appendix Contents}
\phantomsection
\label{sec:appendix}

\newcommand{\appendixentry}[2]{%
  \noindent\hspace*{#1}\hyperref[#2]{\ref*{#2}\quad\nameref*{#2}}%
  \nobreak\dotfill\hyperref[#2]{\pageref*{#2}}\par}
\begingroup
\small
\setlength{\parskip}{2pt}
\hypersetup{linkcolor=black}
{\bfseries\appendixentry{0pt}{app:mcp_setup}}
\smallskip
{\bfseries\appendixentry{0pt}{app:ptc_bench_construction}}
\appendixentry{1em}{app:simulated_apis}
\appendixentry{1em}{app:task_schema}
\appendixentry{1em}{app:verifier}
\appendixentry{1em}{app:failure_injection}
\appendixentry{1em}{app:validation}
\appendixentry{1em}{app:neutrality}
\smallskip
{\bfseries\appendixentry{0pt}{app:additional}}
\appendixentry{1em}{app:std}
\appendixentry{1em}{app:native}
\appendixentry{1em}{app:scale}
\appendixentry{1em}{app:api_calls}
\endgroup

\section{Evaluation Setup for MCP-Universe and MCP-Bench}
\label{app:mcp_setup}
For MCP-Universe, we report results on the reproducible executable subset available in our setup: 90 tasks from \texttt{financial\_analysis} and \texttt{web\_search}. We choose these suites because they do not require private credentials, paid APIs, or specialized infrastructure, while still covering computation-heavy and retrieval-heavy tool use. For MCP-Bench, we evaluate its reproducible credential-free subset: 70 tasks from 23 free MCP servers. Tasks requiring paid or private credentials are excluded to avoid access-dependent variation unrelated to the tool-use interface. We use MCP-Bench's published Overall Score, the mean of six LLM-judged rubric dimensions normalized to $[0,1]$, so MCP-Bench results are continuous scores rather than pass rates.


\section{PTC-Bench Construction and Validation}
\label{app:ptc_bench_construction}

This appendix section provides additional details on the construction and validation of \ptcb{}. Since \ptcb{} is designed to evaluate long-horizon, production-style tool-use workflows, it naturally stresses properties such as large initial environments, dependent API calls, multi-turn refinement, and transient failures. We describe how tasks, APIs, verifiers, and failure-injection rules are constructed, and clarify why the benchmark is not tied to \iptc{}-specific mechanisms.

\subsection{Simulated APIs}
\label{app:simulated_apis}

\ptcb{} runs on deterministic local API backends. The benchmark uses ten simulated APIs: \texttt{EnterpriseCalendarAPI}, \texttt{PersonalCalendarAPI}, \texttt{OrgChartAPI}, \texttt{YelpAPI}, \texttt{GoogleMapsAPI}, \texttt{UberAPI}, \texttt{DoorDashAPI}, \texttt{GroceryDeliveryAPI}, \texttt{SpotifyAPI}, and \texttt{VenmoAPI}. These APIs cover workflow domains such as calendar scheduling, organizational analysis, restaurant search, route estimation, ride booking, delivery optimization, playlist management, and peer-to-peer payments.

The APIs are deterministic: given the same initial state and call sequence, they return the same results. Some APIs also include state-mutating methods, such as booking a reservation, saving a business, modifying a playlist, or sending a payment. When a task requires state changes, the final backend state is passed to the verifier.

\subsection{Task Schema}
\label{app:task_schema}

Each task is represented as a JSON object containing the user prompt, exposed APIs, initial simulated state, and verifier type. A simplified example is shown below.

\begin{verbatim}
{
  "id": "ptc_org_chart_largest_manager_0",
  "question": [[
    {
      "role": "user",
      "content": "Look through the org chart history and tell me
      which manager has supervised the largest number of different
      employees overall..."
    }
  ]],
  "involved_classes": ["OrgChartAPI"],
  "initial_config": {
    "OrgChartAPI": {
      "employees": [...],
      "reporting_history": [...]
    }
  },
  "verifier_type": "exact_json_match"
}
\end{verbatim}

The field \texttt{id} uniquely identifies the task. The field \texttt{question} stores one or more dialogue turns; single-turn tasks contain one turn, while multi-turn tasks contain multiple turns. The field \texttt{involved\_classes} specifies which APIs are exposed to the agent. The field \texttt{initial\_config} contains the full initial state of the simulated backend. The field \texttt{verifier\_type} selects the verifier used to score the final answer and, when relevant, the final backend state. Error-recovery tasks additionally include a \texttt{flakiness\_config}, described below.

\subsection{Verifier Design}
\label{app:verifier}

Each task is paired with a deterministic verifier. Verifiers evaluate the model's final answer and, when applicable, the final simulated backend state. They do not inspect the agent's reasoning trace, code, variables, cell history, notebook state, or intermediate tool-call sequence. This ensures that models are evaluated by task outcome rather than by whether their internal execution process matches a particular interface.

We use two main verifier styles.

\paragraph{Exact JSON matching.}
For tasks with a fully specified output schema, the verifier extracts JSON from the model response and checks each required field against the ground truth. The extractor is tolerant of fenced code blocks and surrounding prose, but the extracted object must satisfy the requested schema and values.

\begin{verbatim}
def verify(final_answer, final_state, initial_config, ground_truth):
    answer = extract_json_from_text(final_answer)
    if answer is None:
        return VerificationResult(passed=False, score=0.0)

    checks = {
        key: answer.get(key) == expected
        for key, expected in ground_truth.items()
    }

    passed, score = score_from_checks(checks)
    return VerificationResult(
        passed=passed,
        score=score,
        checks=checks
    )
\end{verbatim}

\paragraph{Recompute-from-state verification.}
For tasks whose expected answer depends on the simulated environment, the verifier recomputes the expected result directly from \texttt{initial\_config}. For example, a restaurant-search verifier may recompute which businesses satisfy cuisine, distance, review-count, open-status, and wait-time constraints, then compare the model's answer by result count, entity set, ordering, and per-field values. Numeric fields use explicit tolerances when appropriate, such as small tolerances for distance or rating comparisons.

This verifier design gives credit to any model that produces the correct final answer or final backend state, regardless of whether it used direct tool calling, stateless \ptc{}, \iptc{}, or provider-native code execution.

\subsection{Failure Injection}
\label{app:failure_injection}

The error-recovery subset uses deterministic transient failures rather than random failures. Each affected task specifies which API method fails and at which call index. For example:

\begin{verbatim}
"flakiness_config": {
  "VenmoAPI": {
    "get_transaction_history": 1
  }
}
\end{verbatim}

This means that the first call to \texttt{get\_transaction\_history} raises a retryable internal error, while the next call succeeds. The failure is deterministic, recoverable, and local to the task instance. It does not persist across tasks.

From the model's perspective, a flaky API is not presented differently from a normal API. The prompt does not mention that a failure will occur, the API documentation does not reveal that the backend is flaky, and the model does not receive any special hint that a task belongs to the error-recovery subset. The only signal available to the model is the runtime error returned by the API call itself, which follows the same structured exception format as other API errors. Thus, the model must infer whether retrying is appropriate from the observed error, rather than from benchmark-specific metadata.

The purpose of this subset is to evaluate whether an agent can recover from production-style transient failures, not whether it can handle permanently broken APIs. Any agent interface with reliable retry behavior can benefit from this subset; the mechanism is not specific to \iptc{}.

\subsection{Annotation and Validation Process}
\label{app:validation}

Tasks were constructed and validated through a multi-stage process. First, we manually authored natural-language queries to reflect production-level user expectations and realistic daily workflow needs. We then constructed the corresponding simulated initial states and expected answers, ensuring that each query uniquely specified the intended result, including tie-breakers and output format when necessary.

Second, we implemented a verifier for each task with respect to its expected answer. The verifier either checks a fixed expected JSON object or recomputes the expected result from the initial simulated state. This keeps scoring independent of the agent's trace.

Third, every task was manually reviewed by two authors. Reviewers checked that the prompt was unambiguous, the initial state was internally consistent, the verifier matched the prompt, and the expected answer did not depend on unspecified assumptions such as missing tie-breakers, hidden time-zone conventions, or ambiguous default behavior. Disagreements were resolved by editing the prompt, state, or verifier until both reviewers agreed.

Fourth, tasks were executed end-to-end with strong baseline models to identify verifier bugs, impossible tasks, or tasks that were trivially solved under all interaction modes. Such tasks were revised before inclusion. For the error-recovery subset, failure locations were selected so that the injected failure occurs on a method required by a correct solution and can be recovered by retrying.

\subsection{Neutrality with Respect to \iptc{}}
\label{app:neutrality}

We acknowledge that \ptcb{} is motivated by the same limitations that \iptc{} addresses: long-horizon workflows, large intermediate states, repeated API calls, multi-turn reuse, and transient failures. However, the benchmark is not defined in terms of \iptc{}-specific mechanisms.

First, the task prompts do not mention Python, notebooks, persistent cells, variables, or code execution. The tasks are written as natural-language workflow requests over simulated APIs.

Second, all interaction modes are evaluated on the same prompts, same initial states, same API implementations, and same verifiers. The only difference across modes is the interface through which the model interacts with the tools.

Third, verifiers do not inspect \iptc{}-specific artifacts. They evaluate only the final answer, final backend state, initial configuration, and ground truth.

Fourth, the multi-turn and error-recovery subsets reward general agent capabilities rather than a specific implementation. Multi-turn tasks benefit any agent that can preserve or reconstruct useful intermediate information, including memory, caching, retrieval, or explicit recomputation. Error-recovery tasks benefit any agent that can recognize a retryable transient failure and issue a successful retry.

Therefore, \ptcb{} should be understood as a benchmark for long-horizon, stateful, and failure-prone tool use. \iptc{} is one interface designed for this setting, but the benchmark itself can evaluate any tool-use system that interacts with the same APIs and verifiers.

\section{Additional Experimental Results}
\label{app:additional}

This section reports additional experiments. Unless noted otherwise, results are averaged over three independent end-to-end runs on the same task sets, and $\pm$ denotes the standard deviation across the three run-level averages.

\subsection{Run-to-Run Variation in the Main Results}
\label{app:std}

Table~\ref{tab:app_std} reports the standard deviation of every entry in Table~\ref{tab:main_results_wip}. Each run's accuracy and per-task token averages are pooled over all 240 tasks, and the standard deviation is taken across the three run-level values. 

\ifdefined\arxivlayout
\begin{table}[!htbp]
\else
\begin{table}[h]
\fi
\centering
\scriptsize
\caption{\textbf{Standard deviation across three runs} for each entry of Table~\ref{tab:main_results_wip}. The last row averages the per-model standard deviations.}
\label{tab:app_std}
\vspace{0.5em}
\begin{adjustbox}{max width=\textwidth}
\begin{tabular}{lccc ccc ccc}
\toprule
\multirow{2}{*}{Model}
& \multicolumn{3}{c}{Accuracy}
& \multicolumn{3}{c}{Raw Tokens}
& \multicolumn{3}{c}{Trajectory Tokens} \\
\cmidrule(lr){2-4}
\cmidrule(lr){5-7}
\cmidrule(lr){8-10}
& Tool-only & PTC & I-PTC
& Tool-only & PTC & I-PTC
& Tool-only & PTC & I-PTC \\
\midrule
GPT 5.4
& 0.2 & 1.8 & 2.5
& 1.6K & 0.7K & 0.7K
& 0.13K & 0.31K & 0.12K \\
GPT 5.4 Mini
& 0.5 & 1.7 & 1.8
& 2.4K & 2.1K & 0.4K
& 0.13K & 0.08K & 0.08K \\
GPT 5.4 Nano
& 0.2 & 2.2 & 1.0
& 2.6K & 2.5K & 4.4K
& 0.25K & 0.12K & 0.36K \\
\midrule
Gemini 3.1 Pro
& 2.2 & 0.6 & 0.4
& 5.0K & 3.3K & 4.0K
& 0.16K & 0.50K & 0.35K \\
Gemini 3.1 Flash-Lite
& 0.8 & 2.1 & 1.1
& 6.4K & 6.4K & 5.0K
& 0.78K & 0.42K & 0.43K \\
Gemini 2.5 Pro
& 2.8 & 0.1 & 0.4
& 4.5K & 2.6K & 0.6K
& 0.23K & 0.18K & 0.07K \\
\midrule
Claude Opus 4.7
& 7.1 & 1.4 & 1.5
& 26.9K & 2.1K & 2.5K
& 1.89K & 0.11K & 0.05K \\
Claude Sonnet 4.6
& 1.0 & 0.6 & 0.5
& 2.2K & 1.1K & 2.9K
& 0.16K & 0.32K & 0.05K \\
\midrule
Qwen 3.5-397B-A17B
& 0.6 & 0.9 & 0.6
& 5.9K & 7.3K & 3.3K
& 0.19K & 0.46K & 0.24K \\
Kimi K2.5$^\dagger$
& 0.9 & 2.6 & 2.2
& 4.1K & 4.1K & 3.6K
& 0.16K & 0.27K & 0.24K \\
\midrule
Mean (10 models)
& 1.6 & 1.4 & 1.2
& 6.2K & 3.2K & 2.7K
& 0.41K & 0.28K & 0.20K \\
\bottomrule
\end{tabular}
\end{adjustbox}
\end{table}

\subsection{Provider-Native Code Execution}
\label{app:native}

Besides Anthropic's built-in code execution tool, we evaluate provider-native code execution for GPT and Gemini, in which model-written code calls the benchmark APIs directly. For GPT, we use Codex's code-only mode, an experimental feature at the time of our experiments (July 2026): the model can only write JavaScript, which runs in a fresh sandbox per invocation, and the benchmark APIs are exposed as MCP servers callable only from that code. At that time, Gemini offered no equivalent runtime. Gemini CLI (v0.53.0) supported tail tool calls, in which a tool's result can trigger another tool call whose result replaces it, but no interpreter used them, so we implemented a minimal Python interpreter on this mechanism through the CLI's documented hook interface. In both cases, intermediate API results stay inside the code rather than entering the model's context. We replace the provider's system prompt with our \texttt{native\_ptc} instruction, disable shell and web tools, reject API calls issued directly by the model so that the APIs are reachable only from code, keep no state between invocations, and score with the same verifiers. An audit of all traces found no access to scenario files, answer keys, or credentials. The GPT and Gemini results are single runs over all 240 tasks, since each run requires a full CLI harness; Claude results come from the runs of Table~\ref{tab:main_results_wip}.

Provider-native code execution outperforms traditional tool calling on every model (Table~\ref{tab:app_native}). \iptc{} is more accurate than it on GPT-5.4 and on all three Gemini models, by 7.5--13.8 points on Gemini, and matches it on both Claude models ($82.9\pm1.5$ vs.\ $82.6\pm0.5$ on Opus 4.7, $80.6\pm0.5$ vs.\ $80.0\pm4.1$ on Sonnet 4.6). Codex exceeds the Table~\ref{tab:main_results_wip} \iptc{} results on GPT-5.4 mini and nano; in a matched single run on GPT-5.4 mini with \iptc{}'s documentation learner disabled, the two are statistically tied ($73.6\%$ vs.\ $74.2\%$, exact McNemar $p=1.0$).

Provider-native code execution is also more expensive (Table~\ref{tab:app_native_cost}), using $2.5$--$7.5\times$ more raw tokens, $2.9$--$4.8\times$ more trajectory tokens, and, where measured, $3.6$--$3.9\times$ more wall-clock time. The overhead comes from orchestration, not computation: each Gemini API call is a full CLI round trip (115--176 hook process launches per task, against at most 0.4\,s of interpreter execution), and Codex re-sends its harness context on every model turn, whereas \iptc{} runs the program inside one persistent environment.

The GPT comparison is not compute-matched. Codex runs on the Responses API with reasoning enabled and used about 2.4K reasoning tokens per task, whereas the chat-completions function-tool endpoint used for our GPT runs did not allow reasoning alongside function tools. The GPT-5.4 mini and nano differences may therefore partly reflect unequal inference budgets rather than the interface.

For completeness, Table~\ref{tab:app_native} also reports a second kind of provider code execution, which we call hosted interpreters: OpenAI's \texttt{code\_interpreter} tool and Gemini's \texttt{code\_execution} tool. A hosted interpreter runs model-written code on the provider's servers, but that code cannot call any tools. The model must therefore still call each benchmark API itself, one tool call at a time, and can use the interpreter only to compute over results it has already received, for example by copying them into the code. This is traditional tool calling with an added code sandbox for computation, not \ptc{}, where the code itself calls the tools and intermediate results never reach the model. Averaged over three runs, hosted interpreters stay near traditional tool calling on the GPT models and fall below it on Gemini 3.1 Pro ($51.7\%$ vs.\ $67.9\%$), helping only on Gemini 3.1 Flash-Lite ($39.4\%$ vs.\ $10.8\%$).

\ifdefined\arxivlayout
\begin{table}[!htbp]
\centering
\footnotesize
\else
\begin{table}[h]
\centering
\scriptsize
\begin{minipage}[t]{0.52\textwidth}
\fi
\centering
\caption{\textbf{Provider-native code execution} (pass rate, \%). Provider-native: Codex code-only mode (GPT) and our interpreter on Gemini CLI tail tool calls (Gemini), single runs; Anthropic's built-in code execution tool (Claude), from the runs of Table~\ref{tab:main_results_wip}. Hosted interpreters: mean $\pm$ std over three runs. \iptc{}: Table~\ref{tab:main_results_wip}.}
\label{tab:app_native}
\begin{tabular}{lccc}
\toprule
Model & Hosted interp. & Provider-native & \iptc{} \\
\midrule
GPT-5.4 & $17.6\pm3.6$ & 76.2 & \textbf{80.0} \\
GPT-5.4 mini & $9.6\pm1.1$ & \textbf{74.2} & 68.0 \\
GPT-5.4 nano & $6.7\pm0.0$ & \textbf{48.3} & 46.4 \\
Gemini 3.1 Pro & $51.7\pm1.2$ & 75.4 & \textbf{82.9} \\
Gemini 3.1 Flash-Lite & $39.4\pm3.2$ & 60.8$^{\dagger}$ & \textbf{72.8} \\
Gemini 2.5 Pro & N/A$^{*}$ & 46.2 & \textbf{60.0} \\
Claude Opus 4.7 & N/A & 82.6 & \textbf{82.9} \\
Claude Sonnet 4.6 & N/A & 80.0 & \textbf{80.6} \\
\bottomrule
\end{tabular}

\vspace{2pt}
{\ifdefined\arxivlayout\footnotesize\raggedright\else\tiny\fi $^{*}$In July 2026, the Gemini API rejected the combined interpreter and function-tool configuration for Gemini 2.5 (``Tool call context circulation is not enabled''). $^{\dagger}$Run with \texttt{gemini-3.1-flash-lite}; other Flash-Lite entries use \texttt{gemini-3.1-flash-lite-preview}.\ifdefined\arxivlayout\par\fi}
\ifdefined\arxivlayout
\end{table}
\begin{table}[!htbp]
\centering
\footnotesize
\else
\end{minipage}
\hfill
\begin{minipage}[t]{0.44\textwidth}
\centering
\fi
\caption{\textbf{Cost of provider-native code execution} vs.\ \iptc{} (per task). \iptc{} rows for GPT-5.4 mini and Gemini 3.1 Pro are matched single runs with the documentation learner disabled (73.6\% and 82.5\% pass). Claude rows come from the runs of Table~\ref{tab:main_results_wip}, which did not record wall-clock time.}
\label{tab:app_native_cost}
\begin{tabular}{lccc}
\toprule
Runtime & Raw & Traj. & Wall \\
\midrule
GPT-5.4 mini, Codex & 332K & 22.4K & 55.5\,s \\
GPT-5.4 mini, \iptc{} & \textbf{44K} & \textbf{7.6K} & \textbf{15.5\,s} \\
\midrule
Gemini 3.1 Pro, Gemini CLI & 174K & 22.9K & 211.2\,s \\
Gemini 3.1 Pro, \iptc{} & \textbf{61K} & \textbf{7.9K} & \textbf{53.7\,s} \\
\midrule
Claude Opus 4.7, built-in & 246K & 36.2K & -- \\
Claude Opus 4.7, \iptc{} & \textbf{85K} & \textbf{7.6K} & -- \\
\midrule
Claude Sonnet 4.6, built-in & 280K & 35.6K & -- \\
Claude Sonnet 4.6, \iptc{} & \textbf{112K} & \textbf{10.7K} & -- \\
\bottomrule
\end{tabular}
\ifdefined\arxivlayout\else
\end{minipage}
\fi
\end{table}

\subsection{Model Capability and Scale}
\label{app:scale}

Programmatic interaction requires sufficient instruction following, code generation, and tracking of execution state. Table~\ref{tab:app_scale} evaluates the Qwen3.5 family from 0.8B to 397B total parameters on all 240 \ptcb{} tasks, comparing traditional tool calling, one-shot \ptc{}, and raw \iptc{}. At 0.8B and 2B, both programmatic interfaces slightly underperform traditional tool calling, suggesting that these models cannot yet reliably generate and execute multi-step tool-use programs. Programmatic calling first becomes beneficial at 4B, where raw \iptc{} improves pass rate from 26.9\% to 32.8\%, and for every larger model at least one programmatic interface improves substantially over tool calling. In practice, smaller models may benefit from constrained code templates or adaptive routing among tool calling, \ptc{}, and \iptc{}.

\ifdefined\arxivlayout
\begin{table}[!htbp]
\else
\begin{table}[h]
\fi
\centering
\scriptsize
\caption{\textbf{Qwen3.5 model-scale study} (pass rate, \%, all 240 \ptcb{} tasks). Raw \iptc{} uses persistent cells without inspection or documentation updates.}
\label{tab:app_scale}
\vspace{0.5em}
\begin{tabular}{lcccc}
\toprule
Model & Active params & Tool-only & \ptc{} & Raw \iptc{} \\
\midrule
Qwen3.5-0.8B & 0.8B & $\mathbf{5.3\pm1.0}$ & $4.6\pm0.4$ & $4.4\pm1.0$ \\
Qwen3.5-2B & 2B & $\mathbf{10.3\pm1.0}$ & $8.9\pm1.2$ & $8.7\pm1.7$ \\
Qwen3.5-4B & 4B & $26.9\pm1.7$ & $32.2\pm0.9$ & $\mathbf{32.8\pm3.0}$ \\
Qwen3.5-9B & 9B & $32.1\pm1.0$ & $43.5\pm1.7$ & $\mathbf{46.8\pm1.0}$ \\
Qwen3.5-27B & 27B & $62.6\pm2.1$ & $\mathbf{79.4\pm0.2}$ & $78.6\pm1.3$ \\
Qwen3.5-35B-A3B & 3B & $36.7\pm1.8$ & $66.2\pm1.7$ & $\mathbf{67.9\pm0.8}$ \\
Qwen3.5-122B-A10B & 10B & $59.6\pm0.8$ & $72.2\pm0.2$ & $\mathbf{74.4\pm2.1}$ \\
Qwen3.5-397B-A17B & 17B & $43.5\pm0.6$ & $75.3\pm0.9$ & $\mathbf{77.8\pm1.0}$ \\
\bottomrule
\end{tabular}
\end{table}

\subsection{Backend API Calls}
\label{app:api_calls}

Table~\ref{tab:app_api_calls} reports the mean number of backend API calls per task. Relative to traditional tool calling, programmatic interfaces save tokens because loops, filtering, joins, and aggregation run inside Python: one program or cell can invoke many backend APIs in a single model round without returning every intermediate result to the context. Relative to \ptc{}, \iptc{}'s further token savings come mainly from on-demand inspection (Table~\ref{tab:component_ablation}), while persistent state reduces redundant backend calls: raw \iptc{} makes 19--40\% fewer calls than stateless \ptc{}. Tool-only makes the fewest calls, but this reflects early stopping rather than efficiency: for these models it solves only 10--19\% of tasks, versus 59--80\% for the programmatic interfaces.

\ifdefined\arxivlayout
\begin{table}[!htbp]
\else
\begin{table}[h]
\fi
\centering
\scriptsize
\caption{\textbf{Mean backend API calls per task} on all 240 \ptcb{} tasks.}
\label{tab:app_api_calls}
\vspace{0.5em}
\begin{tabular}{lccc}
\toprule
Model & Tool-only & \ptc{} & Raw \iptc{} \\
\midrule
GPT-5.4 & $30.8\pm1.4$ & $83.0\pm7.4$ & $67.1\pm5.2$ \\
GPT-5.4 mini & $45.0\pm1.8$ & $135.1\pm32.3$ & $81.6\pm7.3$ \\
Gemini 3.1 Flash-Lite & $26.2\pm1.5$ & $114.9\pm8.1$ & $84.8\pm7.2$ \\
\bottomrule
\end{tabular}
\end{table}

\clearpage 

\end{document}